\documentclass[10pt,twocolumn,letterpaper]{article}

\usepackage{iccv}
\usepackage[breaklinks=true,bookmarks=false]{hyperref}
\usepackage{graphicx}
\usepackage{xspace}
\usepackage{tabularx}
\usepackage{booktabs}
\usepackage{amsmath}
\usepackage{amssymb}
\usepackage{xcolor}
\usepackage{times}
\usepackage{epsfig}
\usepackage{paralist}
\usepackage[numbers,compress]{natbib}
\usepackage{subcaption}
\usepackage{multirow}
\usepackage{makecell}
\usepackage{tabularx}
\usepackage[capitalize,noabbrev]{cleveref}
\usepackage[accsupp]{axessibility}

\crefname{section}{Sec.}{Secs.}
\crefname{table}{Tab.}{Tabs.}
\crefname{figure}{Fig.}{Figs.}
\crefname{appendix}{App.}{Apps.}
\newcommand{\denselist}{\itemsep 0pt\parsep=0pt\partopsep 0pt\vspace{-\topsep}}
\setdefaultleftmargin{1.5em}{}{}{}{.5em}{.5em}

\expandafter\def\expandafter\normalsize\expandafter{%
    \normalsize%
    \setlength\abovedisplayskip{2pt}%
    \setlength\belowdisplayskip{2pt}%
    \setlength\abovedisplayshortskip{-8pt}%
    \setlength\belowdisplayshortskip{2pt}%
}

\newcommand{\xhdr}[1]{\vspace{2pt}\noindent\textbf{#1}\xspace}

\DeclareMathOperator{\IoU}{IoU}
\DeclareMathOperator{\Cost}{Cost}
\DeclareMathOperator{\MLP}{MLP}

\DeclareMathOperator{\conv}{conv}

\newcommand{\R}{\ensuremath{\mathbb{R}}}
\newcommand{\predth}{\ensuremath{\tau_\text{pred}}\xspace}
\newcommand{\ioutrainth}{\ensuremath{\tau_\text{train}}\xspace}
\newcommand{\iouevalth}{\ensuremath{\tau_\text{eval}}\xspace}
\newcommand{\PROJECTNAME}{Multi3DRefer\xspace}
\newcommand{\OURMETHOD}{M3DRef\xspace}
\newcommand{\OURMETHODCLIP}{M3DRef-CLIP\xspace}
\newcommand{\NUMZEROGT}{6688\xspace}
\newcommand{\NUMUNIQUEGT}{42060\xspace}
\newcommand{\NUMMULTIPLEGTS}{13178\xspace}
\newcommand{\NUMTOTAL}{61926\xspace}

\newcolumntype{Y}{>{\centering\arraybackslash}X}

\iccvfinalcopy

\def\iccvPaperID{12551}

\ificcvfinal\fi

\begin{document}

\title{\PROJECTNAME: Grounding Text Description to Multiple 3D Objects}

\author{Yiming Zhang$^1$ \quad ZeMing Gong$^1$ \quad Angel X. Chang$^{1,2}$\\\
Simon Fraser University$^1$ \quad Alberta Machine Intelligence Institute (Amii)$^2$\\
{\tt\small \{yza440, zmgong, angelx\}@sfu.ca} \\
\footnotesize{\url{https://3dlg-hcvc.github.io/multi3drefer/}}
}

\maketitle
\ificcvfinal\thispagestyle{empty}\fi

\begin{abstract}
We introduce the task of localizing a flexible number of objects in real-world 3D scenes using natural language descriptions. Existing 3D visual grounding tasks focus on localizing a unique object given a text description. However, such a strict setting is unnatural as localizing potentially multiple objects is a common need in real-world scenarios and robotic tasks (e.g., visual navigation and object rearrangement). To address this setting we propose \PROJECTNAME, generalizing the ScanRefer dataset and task. Our dataset contains \NUMTOTAL descriptions of 11609 objects, where zero, single or multiple target objects are referenced by each description.
We also introduce a new evaluation metric and benchmark methods from prior work to enable further investigation of multi-modal 3D scene understanding. Furthermore, we develop a better baseline leveraging 2D features from CLIP by rendering object proposals online with contrastive learning, which outperforms the state of the art on the ScanRefer benchmark.
\end{abstract}
\section{Introduction}

There is growing interest in multi-modal methods that connect language and vision, tackling tasks such as image captioning, visual question answering, text-to-image retrieval, and generation. One fundamental task is visual grounding where natural language text queries are linked to regions of an image or 3D scene.
While the problem of visual grounding has been well studied in 2D images, there are fewer datasets and methods studying the problem in 3D scenes.
Being able to indicate the object that the text ``the first of four stools'' references in a 3D scene is useful for applications in robotics, AR/VR, and online 3D environments where we have access to not just a static image, but a 3D scene.
Work on 3D datasets for visual grounding~\cite{chen2020scanrefer,achlioptas2020referit_3d} has spurred the development of methods for 3D visual grounding~\cite{luo20223d,zhao20213dvg,yuan2021instancerefer, huang2022multi,huang2021text,abdelreheem20223dreftransformer,yang2021sat,chen2020scanrefer} and the inverse task of 3D captioning~\cite{chen2020scanrefer}, as well as unified methods that tackle both~\cite{chen2022d3net, cai20223djcg, chen2022unit3d}. 

\begin{figure}[t]
  \includegraphics[width=\linewidth]{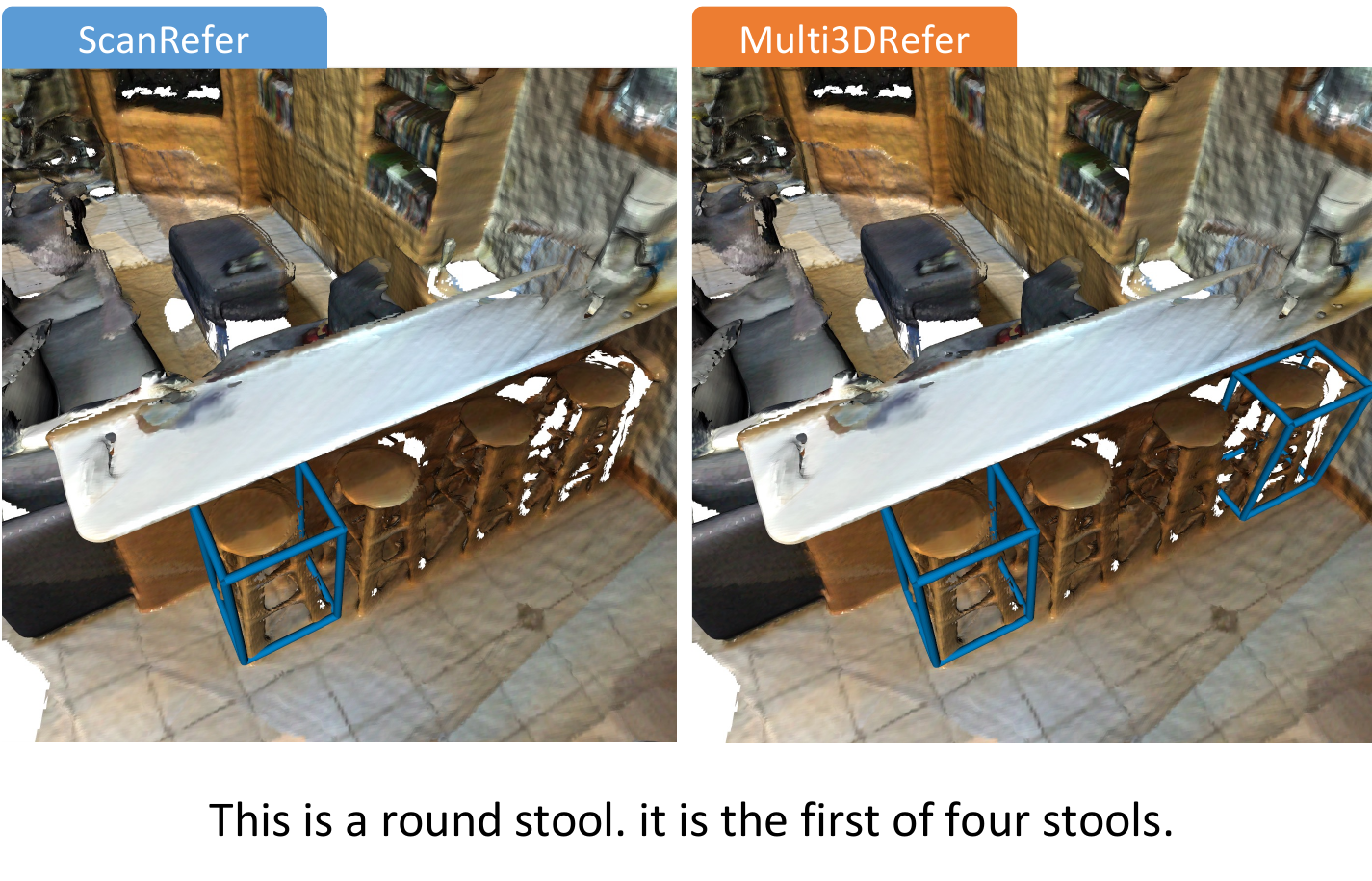}
  \vspace{-22pt}
  \caption{We introduce \PROJECTNAME, a dataset and task where there are potentially multiple target objects for a given description.  
  In ScanRefer~\cite{chen2020scanrefer} (left), the description corresponds to exactly one object (blue box), while in \PROJECTNAME (right), there are multiple target objects.
  }
  \label{fig:intro_comparison}
\end{figure}

However, existing datasets and tasks \cite{chen2020scanrefer, achlioptas2020referit_3d} are designed with the assumption that there is a unique target object when performing visual grounding in a 3D scene. This assumption makes ambiguous descriptions that may refer to multiple objects problematic (see \Cref{fig:intro_comparison}). Furthermore, this explicitly discourages visual grounding methods from demonstrating generalization to similar object instances on the basis of common features of the objects (e.g., similar size, color, texture) and spatial relations between objects (e.g., first in a row left-to-right or right-to-left).

We address these shortcomings with an enhanced dataset and task that we call \PROJECTNAME where a flexible number of target objects (zero, single or multiple) in a 3D scene are localized given language descriptions. We modify and enhance language data from ScanRefer~\cite{chen2020scanrefer} and propose evaluation metrics to benchmark prior work and a CLIP-based~\cite{radford2021learning} method that we propose on the flexible number visual grounding task. In summary, we make the following contributions:
1) generalize 3D visual grounding to a flexible number of target objects given natural language descriptions.
2) create an enhanced dataset based on ScanRefer~\cite{chen2020scanrefer} with augmentations from ChatGPT\footnote{\url{https://openai.com/blog/chatgpt/}}, consisting of \NUMTOTAL descriptions in 800 ScanNet V2~\cite{dai2017scannet} scenes.
3) benchmark three prior 3D visual grounding approaches adapted to \PROJECTNAME
4) design an end-to-end approach leveraging CLIP~\cite{radford2021learning} embeddings and online rendering of object proposals with contrastive learning.

\section{Related work}
In this paper, we focus on description localization where a single description may describe one or more objects in a 3D scene.
Below, we review work in grounding in 2D and 3D, as well as recent work leveraging pre-trained vision-language models for 3D scene understanding.

\xhdr{Visual grounding in 2D.}
A variety of datasets and methods have been proposed to investigate visual grounding tasks such as referring expressions~\cite{kazemzadeh2014referitgame,mao2016generation,hu2016natural} and phrase localization~\cite{plummer2015flickr30k,plummer2017phrase} in 2D images.
These datasets have enabled developing various visual-language grounding models~\cite{yu2018mattnet,zhang2018grounding,yang2019fast,lu2019vilbert,deng2021transvg,li2021referring,zhu2022seqtr}. Typically, in these datasets and tasks each phrase refers to exactly one object. A notable exception is the VGPhraseCut~\cite{wu2020phrasecut} dataset, based on Visual Genome~\cite{krishna2017visual} using templated phrases where each phrase is grounded to potentially multiple instance segments.

Recent work in language and vision has started to tackle more flexible grounding. 
\citet{kim2022flexible} noted that not all queries can be visually grounded (i.e. it is possible to have no targets) and constructed a dataset to study grounding performance when there are unanswerable queries. \citet{kuo2022findit} proposed a single model for referring expression comprehension, object detection, and phrase localization. While their model can handle multiple objects, the queries for multiple objects are typically short and category-based. Recent work~\cite{kamath2021mdetr,li2022grounded} reframed the problem of object detection as phrase grounding by introducing losses to align words to regions. These methods are flexible and have been used to improve both detection and visual grounding. In our work, we construct a dataset for flexible grounding in 3D.

\xhdr{Visual grounding in 3D.}
Early work studied selecting the correct 3D shapes based on a text description in reference game setups~\cite{achlioptas2019shapeglot,thomason2022language,koo2022partglot}, as well as learning joint language-3D embeddings for 3D text-to-shape retrieval~\cite{chen2018text2shape,ruan2022tricolo}.
These works focused on descriptions of single objects in isolation. Moving beyond single 3D objects, researchers also studied grounding of language to objects in 3D scenes.
At the scene level, ScanRefer~\cite{chen2020scanrefer} and ReferIt3D~\cite{achlioptas2020referit_3d} introduce two datasets consisting of language descriptions of 3D objects from the real-world dataset ScanNet~\cite{dai2017scannet}.
In detail, ReferIt3D~\cite{achlioptas2020referit_3d} contains both template-based descriptions generated based on spatial relations between objects (Sr3D) and human-annotated fine-grained descriptions (Nr3D).
They also propose two different grounding tasks, both localizing a unique target object referred by a description.
ScanRefer~\cite{chen2020scanrefer} requires both object detection and grounding, while ReferIt3D~\cite{achlioptas2020referit_3d} focuses on discriminating a target object from multiple objects of that semantic class given ground-truth object bounding boxes.

Different approaches~\cite{chen2020scanrefer,achlioptas2020referit_3d} have been proposed to tackle the two tasks, with models focusing on graph representations~\cite{huang2021text,yuan2021instancerefer,feng2021free}, improved handling of relations~\cite{zhao20213dvg,chen2022language}, neurosymbolic reasoning~\cite{hsu2023ns3d}, leveraging multi-view images and 2D semantics~\cite{yang2021sat,jain2021looking,huang2022multi,bakr2022look}, to unified models that can address both grounding and captioning~\cite{cai20223djcg,chen2022d3net,huang2022unified,chen2022unit3d}.  Recently, \citet{abdelreheem2022scanents3d,wu2023eda} showed that including training on dense annotations can improve performance.
\citet{jain2022bottom} proposed to tackle object detection and visual grounding in a unified way by aligning features for text tokens with object proposals.
In this work, we compare the performance of three recent models on our task \PROJECTNAME with a new CLIP-based~\cite{radford2021learning} model.

\xhdr{3D understanding using vision-language models.}
Large pre-trained text-vision models such as CLIP~\cite{radford2021learning} and ALIGN~\cite{jia2021scaling} enabled work leveraging these models for 3D scene understanding.
Recent work learn joint embeddings with text-image-3D representations~\cite{xue2022ulip,zhang2022pointclip}, used for disambiguating referring expressions~\cite{thomason2022language} or text-to-shape retrieval~\cite{ruan2022tricolo}.
Incorporating pre-trained 2D visual features also enabled expansion of 3D detection and instance segmentation to a larger number of categories~\cite{rozenberszki2022language}, as well as tackling open vocabulary 3D detection~\cite{lu2022open,shafiullah2022clip}, and building of 3D semantic maps~\cite{jatavallabhula2023conceptfusion}.
In our work, we show that we can leverage CLIP~\cite{radford2021learning} for improved visual grounding.

\begin{figure}  
  \includegraphics[width=\linewidth]{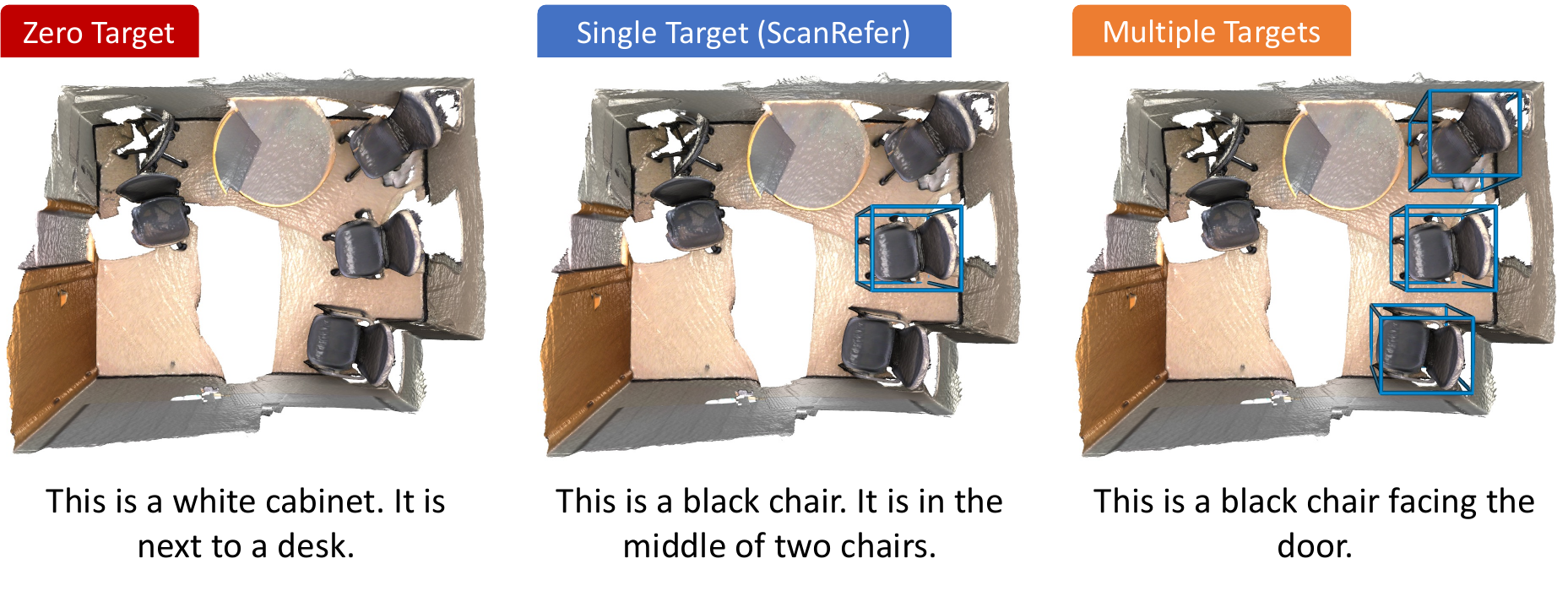}
  \vspace{-20pt}
  \caption{Example description-scene pairs in the \PROJECTNAME dataset with zero, single, or multiple target objects. Blue boxes indicate ground truth target objects.}
  \label{fig:dataset_demo}
\end{figure}

\section{\PROJECTNAME dataset}
\begin{figure}
  \centering  \includegraphics[width=0.9\linewidth]{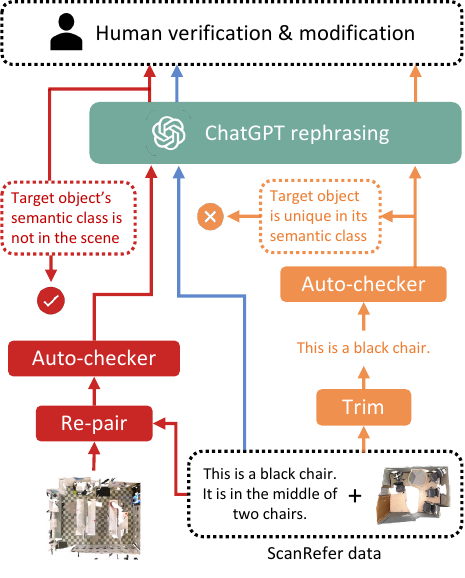}
  \vspace{-8pt}
  \caption{
  \PROJECTNAME data construction pipeline showing the generation process of Zero Target (red), Single Target (blue) and Multiple Target (orange) data.
  We use ChatGPT to add diversity to the description.
  All scene-description pairs are manually verified and modified if needed.
  }
  \label{fig:dataset_construction}
\end{figure}

To study our task, we build the \PROJECTNAME dataset, a superset of the existing ScanRefer dataset~\cite{chen2020scanrefer} with language descriptions of varying granularities.
We augment ScanRefer to create a dataset with 3 types of description-scene pairs: a) Zero Target; b) Single Target; and c) Multiple Targets, indicating zero, single, or multiple target objects in the scene match the description (see \Cref{fig:dataset_demo}).
In addition, we use ChatGPT to augment the descriptions so they are more natural and diverse (see \Cref{fig:dataset_construction} for the overall data construction pipeline).
To ensure the dataset is of high quality we manually verify all generated samples.
We obtain a dataset with \NUMTOTAL descriptions in total (see \Cref{tab:datasets_comparison,tab:data_attribute_breakdown} for statistics).

\begin{table}
\resizebox{\linewidth}{!}
{
\begin{tabular}{@{} l rrrr@{}}
\toprule
Dataset & Zero Target & Single Target & Multiple Targets & \textbf{Total} \\
\midrule
ScanRefer~\cite{chen2020scanrefer} & - & 51583 & - & 51583 \\ %
Sr3D~\cite{achlioptas2020referit_3d} & - & 83572 & - & 83572 \\
Sr3D+~\cite{achlioptas2020referit_3d} & - & 114532 & - & 114532 \\
Nr3D~\cite{achlioptas2020referit_3d} & - & 41503 & - & 41503 \\
\midrule
\PROJECTNAME & \NUMZEROGT & \NUMUNIQUEGT & \NUMMULTIPLEGTS & \NUMTOTAL \\
\bottomrule
\end{tabular}
}
\vspace{-4pt}
\caption{Compared to existing 3D visual grounding datasets, our \PROJECTNAME dataset contains text that describes zero, single, or multiple target objects.}
\label{tab:datasets_comparison}
\end{table}

\subsection{Adapting text for multiple targets}
We start with scene-description pairs from ScanRefer and augment the data to create samples that refer to zero or more targets.
For Single Target descriptions (type b), we take the released ScanRefer dataset which consists of description-scene pairs that have been initially verified to refer to unique objects.
We double-checked whether these descriptions indeed refer to unique objects and we found 9324 descriptions that are ambiguous.
We use these ambiguous descriptions as an initial set of type (c) descriptions that can refer to multiple objects.
For additional type (c) descriptions, we obtain from the ScanRefer authors $7741$ ambiguous descriptions that can refer to multiple objects and annotate those descriptions with matching objects.
To collect type (a) description (no target object) as well as more type (c) descriptions, we develop an efficient data collection pipeline consisting of two stages: 1) automated description generation; and 2) verification and modification.
This pipeline maximizes the use of the existing ScanRefer dataset and reduces manual annotation.
Below, we describe how we generate and verify additional data samples.

\xhdr{Zero Target.}
To obtain descriptions with no target objects, we establish negative pairs of scenes with existing descriptions selected randomly from other scenes.
We then manually verify that descriptions do not match any objects in the new scene.
To reduce the number of description-scene pairs that need to be verified, we automatically check whether the semantic class of the target object for the description appears in the scene.
Only if the semantic class appears in the scene does it need to be manually verified.
Out of 6688 samples, we manually verify 5630 with 1058 automatically checked to have no matching objects.
For pairs that need verification, human annotators are shown the description along with an interactive view of the scene to check that there are no matching objects.

\begin{figure}
  \centering
  \includegraphics[width=0.88\linewidth]{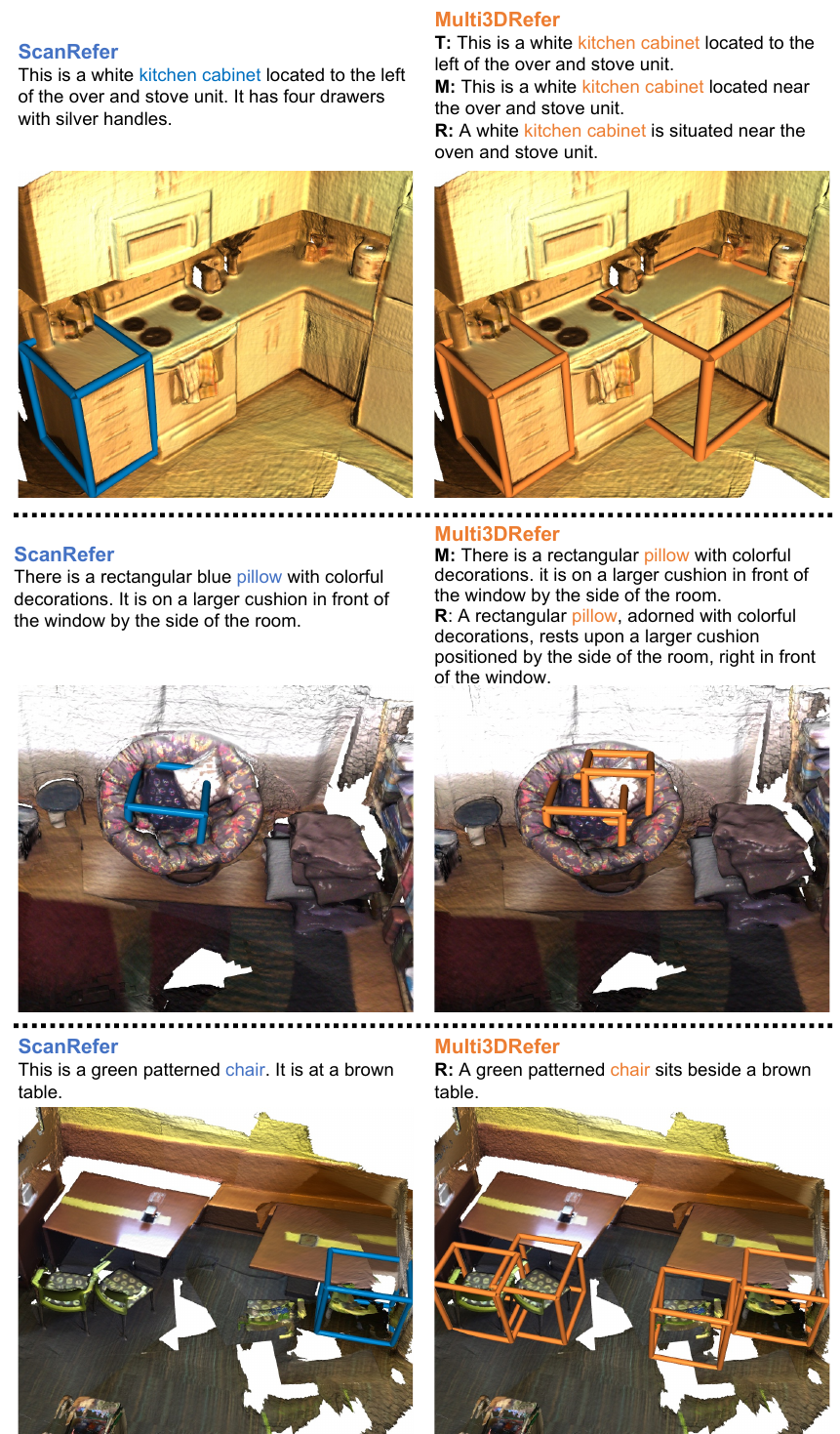}
  \vspace{-4pt}
  \caption{Examples of generated revised descriptions of multiple targets for the \PROJECTNAME dataset where we trim (T) to create more ambiguous descriptions, modify (M) by human, and reword (R) using ChatGPT.
  All descriptions are verified by annotators (see bounding boxes for target objects).
  Note that the reworded (R) descriptions provide more variation in sentence structure.
  }
  \label{fig:mt_demo}
\end{figure}

\xhdr{Multiple Targets.} 
To generate descriptions with multiple targets, we start with the original description-scene pairs in the ScanRefer dataset.
We then randomly select descriptions and trim the text to the first punctuation to obtain shorter, more ambiguous descriptions (e.g., \textit{``The cabinet is white and in the back of the room. It is the one on the left."} $\rightarrow$ \textit{``The cabinet is white and in the back of the room."}). Note that descriptions in which the semantic class of the target object only has a unique object in the scene are skipped.
The trimmed text and scene pairs are sent to the annotation interface.
This time, annotators are asked to select all eligible objects of a description in a 3D interactive scene mesh.
Annotators are also asked to modify trimmed descriptions to fix errors and increase diversity (see \Cref{fig:mt_demo} for examples).

\subsection{Rephrasing using ChatGPT}
To increase description diversity we use the ChatGPT model \emph{text-davinci-002-render} for sentence rephrasing.
We provide ChatGPT with the following prompts:

\begin{compactenum}\denselist
{\footnotesize
\item I will give you a sentence describing an object, please help me polish it and keep its meaning.
\item Help me reword a sentence to a different format but keep its meaning.
\item Help me reword a sentence to make it more natural.
\item Help me reword a sentence describing an object, you should describe the colors and the spatial information in a different way.
\item Help me reword a sentence to an interesting format, you should keep its meaning.
}
\end{compactenum}

\begin{table}
\centering
\resizebox{\linewidth}{!}
{
\begin{tabular}{lrrr}
\toprule
& Unique words & Total words & Avg. description length\\
\midrule
Original & 5067 & 1016190 & 16.4 \\
Rephrased & 7077 & 936935 & 15.1 \\

\bottomrule
\end{tabular}
}
\vspace{-4pt}
\caption{Comparison of \PROJECTNAME language data before and after ChatGPT rephrasing. We count the number of unique words, total words, and the average description length, excluding punctuations.}
\label{tab:data_statistic_for_multi3drefer_original_and_reworded}
\end{table}

\Cref{tab:data_statistic_for_multi3drefer_original_and_reworded} shows statistics comparing the descriptions before and after rephrasing.
We see a richer vocabulary but shorter descriptions after ChatGPT rephrasing.
Below we provide examples of the original (O) and reworded (R) text:
\begin{compactdesc}\denselist
{\footnotesize
\item \textbf{O}: \textit{The table is a round table. It is located between two chairs to the right, and two chairs to the left of it.}
\item \textbf{R}: \textit{The round table is situated between two chairs to its right and two chairs to its left.}
\vspace{4pt}
\item \textbf{O}: \textit{This is a table on the wall in the room. It is next to the window and a few lined-up chairs.}
\item \textbf{R}: \textit{A wall-mounted table resides cozily beside a window in the room, accompanied by a row of orderly chairs.}
\vspace{4pt}
\item \textbf{O}: \textit{A sink on the vanity. It is to the right of the vacuum 
cleaners.}
\item \textbf{R}: \textit{The sink is located on the vanity to the right of the vacuum cleaners.}
\vspace{4pt}
\item \textbf{O}: \textit{This is a white kitchen cabinet located near the over and stove unit.}
\item \textbf{R}: \textit{A white kitchen cabinet is situated near the oven and stove unit}
}
\end{compactdesc}
The rephrased text preserves the original meaning while being more natural.
In addition, ChatGPT automatically corrects typos (e.g., \textit{over} to \textit{oven} in the last example).

\begin{table}
\centering
\resizebox{\linewidth}{!}
{
\begin{tabular}{lrrrrrr}
\toprule
& Spatial & Color & Texture & Shape & \textbf{Total} \\
\midrule
ScanRefer~\cite{chen2020scanrefer} & 51117   & 34692 & 5864  & 17416 & 51583 \\ 
Nr3D~\cite{achlioptas2020referit_3d}& 39711 & 11939 & 526 & 8568 & 41503 \\
Sr3D~\cite{achlioptas2020referit_3d}& 83572 & 6254 & 0 & 648 & 83572 \\
Sr3D+~\cite{achlioptas2020referit_3d}& 114532 & 8666 & 0 & 744 & 114532 \\
\midrule
\PROJECTNAME & 60028 & 41307 & 7121 & 19692 & 61926 \\
\bottomrule
\end{tabular}
}
\vspace{-4pt}
\caption{Breakdown of spatial, color, texture, and shape information in object descriptions from different datasets.}
\label{tab:data_attribute_breakdown}
\end{table}

\subsection{Verification}
After we obtain a set of ChatGPT reworded descriptions, we manually verify the descriptions are well-written and that the object(s) matched in the scene are accurate.
We create a web interface for verifiers to check whether the description matches the identified target objects (see \Cref{sec:supp:web} for details).
The web interface shows the description together with an interactive 3D view of the scene and the target objects.
The verifiers check if the description matches the target objects and only the target objects, or modify the list of target objects (by selecting appropriate objects), or improve the description to fix typos and ambiguities.
The verification was performed by 5 students over a period of one month.

\subsection{Dataset statistics}
In total, our dataset consists of \NUMTOTAL language descriptions, with 51583 directly obtained from ScanRefer, of which $6688$ descriptions match zero-targets and $13178$ match multiple.
For Multiple Targets, the scenes are typically offices or meeting rooms with many chairs and tables. 
See \Cref{tab:datasets_comparison} for a comparison of our final dataset against prior datasets.
We also provide annotations for each description as to whether it refers to spatial, color, texture, or shape attributes (see \Cref{tab:data_attribute_breakdown}).  We provide additional statistics and examples from our dataset in \Cref{sec:supp:data}.

\begin{figure*}
  \includegraphics[width=\linewidth]{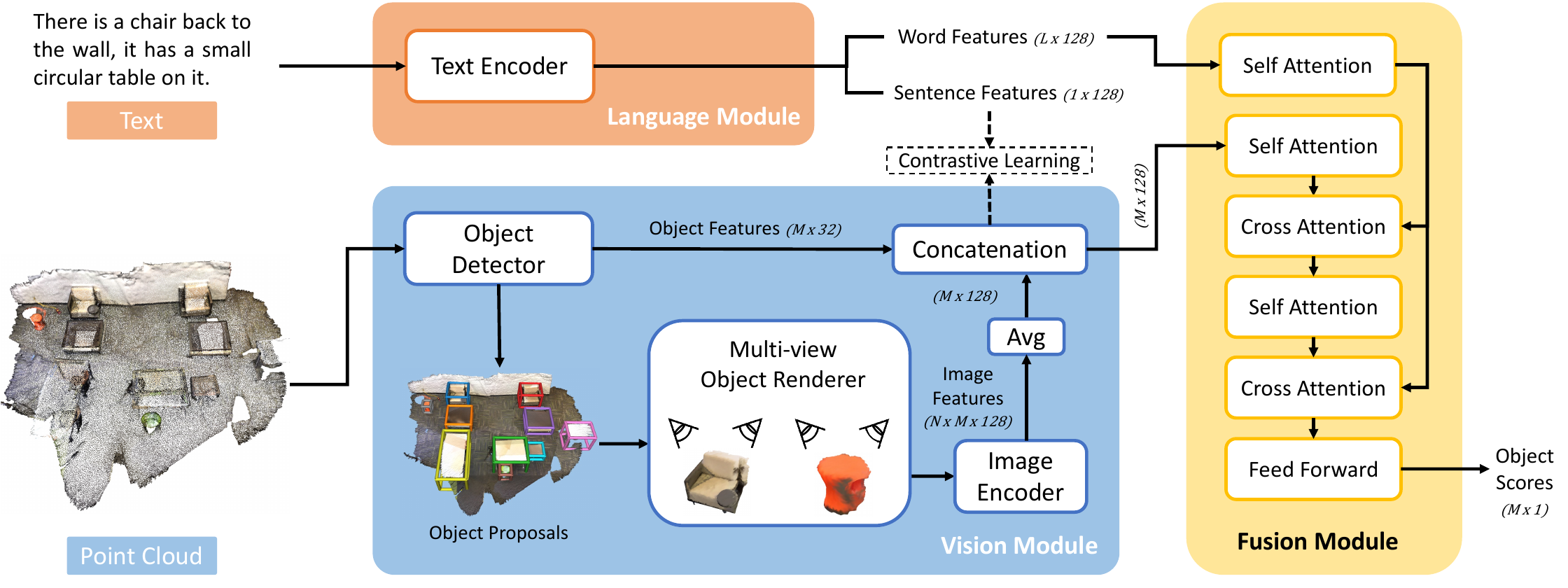}
  \caption{Our M3DRef-CLIP end-to-end architecture.
  Given a scene point cloud and a text description with $L$ tokens (we pad shorter descriptions and truncate longer ones),
  the detector first predicts $M$ object proposals and their 3D features.
  Then an online renderer renders $N$-view images for each proposal and feeds them into the image encoder to get 2D features.
  A transformer-based module then fuses both language features and 2D + 3D object features and outputs scores indicating the match of each object to the description.
  We use PointGroup~\cite{jiang2020pointgroup} to detect and segment the objects in 3D and select CLIP~\cite{radford2021learning} + MLPs as the language and image encoder.
  A contrastive loss is applied between sentence features and object features.
  }
  \label{fig:model_arch}
\end{figure*}

\section{Task}

In the \PROJECTNAME task, we are given as input a 3D real-world scene represented as a point cloud $P \in \mathbb{R}^{N \times (3+C)}$ and a free-form language description with a variable number $M \in \mathbb{N}$ of referred objects, where $N$, $C$ are the number of points and the number of point feature channels, respectively.
The goal is to predict axis-aligned bounding boxes for all $M$ objects that match the description.
Compared to prior work, the difficulty of our task is that the number of referred objects is flexible, i.e., a description can refer to not only one or multiple target objects but also no objects.
Predicting too many or too few target objects are both penalized by our evaluation metrics.

\xhdr{Evaluation metrics.}
To evaluate grounding for a flexible number of target objects, we measure the F1 score at the intersection over union (IoU) thresholds of $\iouevalth=0.25$ and 0.5 (F1@0.25 and F1@0.5).
To investigate model performance for different scenarios, we consider the following 5 cases: a) zero target w/o distractors of the same semantic class; b) zero target w/ distractors; c) single target w/o distractors; d) single target w/ distractors; and e) multiple targets.
Note that c) and d) correspond to the ``unique'' and ``multiple'' cases in ScanRefer~\cite{chen2020scanrefer}.
In addition, a) and c) are easier cases where there are zero or a unique target object of its semantic class in a scene, while b) and d) are more difficult cases containing one or multiple target objects of the same semantic class in a scene.
We also report the micro-average of the 5 cases as an overall score.

During evaluations, we first calculate per-pair IoUs between ground truth and predicted bounding boxes in a scene, and then apply the Hungarian algorithm~\cite{kuhn1955hungarian} to get an optimal one-to-one matching between predicted and GT bounding boxes.
To get the maximum matched IoU, we use the following cost function:
\begin{equation*}
\Cost(i,j) = -\IoU(i,j) \text{ for } {i,j} \in \{1 \ldots N\}
\end{equation*}
where $N=\max(\#\textit{Predictions}, \#\textit{GTs})$.
After obtaining the optimal match, we take pairs with IoUs higher than \iouevalth to be True Positives (TP).
We treat the Zero Target (ZT) case as a special case where recall is always set to $1$, and precision is set to $1$ if there is no prediction or $0$ otherwise.

\section{Method}

We propose \OURMETHODCLIP, a CLIP-based~\cite{radford2021learning} approach and compare to three recent approaches: 3DVG-Transformer~\cite{zhao20213dvg}, 3DJCG~\cite{cai20223djcg} and D3Net~\cite{chen2022d3net}. 
We selected these methods as they were among the top performers on the ScanRefer~\cite{chen2020scanrefer} benchmark\footnote{\href{https://kaldir.vc.in.tum.de/scanrefer_benchmark/}{kaldir.vc.in.tum.de/scanrefer\_benchmark}} with available open-source code.
Note that 3DJCG and D3Net are unified models, which can do both grounding and captioning tasks.
All four models are two-stage approaches, where a 3D object detector first identifies a set of bounding box candidates, and a disambiguation module then selects the target bounding box.
In the original ScanRefer setting, the models are trained using a cross-entropy loss $L_\text{ref}$ where the predicted bounding box has the highest IoU ($>\ioutrainth$) with the GT bounding box as the target bounding box to calculate the loss. 
For our task, we use a binary cross-entropy loss for $L_\text{ref}$ as our problem is a multi-label task (vs classification).

\subsection{\OURMETHODCLIP}

\OURMETHODCLIP follows the two-stage architecture with PointGroup~\cite{jiang2020pointgroup} as the detector, CLIP~\cite{radford2021learning} as the text encoder and a transformer-based fusion module (see \Cref{fig:model_arch}).
We use PointGroup to obtain object proposals along with their 3D features $\mathbf{F}^\text{3d} \in R^{32}$.
The output object proposals are fed into an online object renderer, which renders multi-view 2D images for each object proposal.
We use CLIP to encode the images and use an MLP to project the average of the output 2D image features to obtain $\mathbf{F}^\text{2d} \in \R^{128}$.
The 2D and 3D features are concatenated to form and passed through a 1D convolution (that projects the combined 160-dim features down to 128-dim) to obtain the final visual features $\mathbf{F}^\text{obj} \in \R^{128}$ for an object proposal:
\begin{equation*}
\mathbf{F}^\text{obj} = \conv_\text{1d}([\mathbf{F}^\text{3d}; \mathbf{F}^\text{2d}]),\quad
\mathbf{F}^\text{2d} = \frac{\MLP(\sum_{i=1}^{N}\mathbf{F}_{i}^\text{2d})}{N}
\end{equation*}
We use the CLIP text encoder with an MLP to obtain both 128-dim token-level and sentence embeddings.
We use a transformer-based fusion module to combine object features and token-level embeddings and output confidence scores for each proposal.
To improve the training, we apply a contrastive loss between sentence and object features.

\subsection{Loss function}

We train the network end-to-end with the total loss $L = L_\text{det} + L_{c} + L_\text{ref}$ consisting of the detection loss $L_\text{det}$, contrastive loss $L_\text{c}$, and the reference loss $L_\text{ref}$.
The detection loss $L_\text{det}$ is introduced in PointGroup and consists of four parts: 1) cross-entropy loss for supervising per-point semantic class prediction; 2) $L_1$ loss for supervising per-point offset vector towards object centers; 3) directional loss formed as a mean of minus cosine similarities for further constraining the direction of per-point offset vectors; 4) binary cross-entropy loss for supervising per-point objectness confidence score.
To help learn better multi-modal embeddings, we introduce a symmetric contrastive loss $L_{c}$, 
which can handle a flexible number of target objects:
\begin{equation*}
L_\text{c}^{O\rightarrow S} = -\log\frac{\exp(\cos(\bar{O}_i,S_i)/\tau)}{\sum_{i=1}^{n}\exp(\cos(\bar{O}_i,S_j)/\tau)}
\end{equation*}
\vspace{5pt}
\begin{equation*}
L_\text{c}^{S\rightarrow O} = -\log\frac{\exp(\cos(S_i,\bar{O}_i)/\tau)}{\sum_{i=1}^{n}\exp(\cos(S_i,\bar{O}_j)/\tau)}
\end{equation*}
where $S$ is sentence features, $\bar{O}$ is the mean of object features of all target objects paired with a description, and $\tau$ is a temperature parameter.
Finally, the reference loss $L_\text{ref}$ supervises the matching module to select objects satisfying the description.
We use the multi-class cross-entropy loss as $L_\text{ref}$ for experiments on ScanRefer, Nr3D~\cite{achlioptas2020referit_3d} and \PROJECTNAME (ST) where each description only refers to a single target object.
For experiments on \PROJECTNAME, $L_\text{ref}$ is a sum of binary cross-entropy losses over detected objects.

\xhdr{Training.}
To set up positive (i.e. valid target bounding boxes) and negative instances between the GT bounding boxes and proposed bounding boxes, we use two strategies (see \Cref{sec:supp:analysis} for detailed comparisons):

\xhdr{\textit{All}}
All predicted bounding boxes with IoU higher than the threshold \ioutrainth with GT bounding boxes are considered target bounding boxes.

\xhdr{\textit{Hungarian}}
We apply the Hungarian algorithm~\cite{kuhn1955hungarian} to do bipartite matching, which ensures an optimal solution.
We use the same IoU threshold \ioutrainth to filter prediction-GT bounding box pairs.

\xhdr{Inference.}
At inference time, we take all bounding boxes with predicted scores above the threshold \predth as positives (predicted target bounding boxes for the description).
See \Cref{sec:supp:analysis} for comparisons of different \predth.

\section{Experiments}

\begin{table*}[ht]
\centering
\resizebox{\linewidth}{!}
{
\begin{tabular}{@{}lrrrr rrrr rrrr rrrr rrrr@{}}
\toprule
& \multicolumn{4}{c}{ZT w/o Distractors} & \multicolumn{4}{c}{ZT w Distractors} & \multicolumn{4}{c}{ST w/o Distractors} & \multicolumn{4}{c}{ST w/ Distractors} & \multicolumn{4}{c}{Multiple Targets} \\
\cmidrule(lr){2-5}\cmidrule(lr){6-9}\cmidrule(lr){10-13}\cmidrule(lr){14-17}\cmidrule(lr){18-21}
& Train & Val & Test & \textbf{Total} & Train & Val & Test & \textbf{Total} & Train & Val & Test & \textbf{Total} & Train & Val & Test & \textbf{Total}& Train & Val & Test & \textbf{Total} \\
\midrule
ScanRefer~\cite{chen2020scanrefer} & - & - & - & - & - & - & - & - & 8500 & 2297 & 1201 & 11998 & 28165 & 7211 & 4209 & 39585 & - & -  & - & -\\
\PROJECTNAME & 2702 & 528 & 596 & 3826 & 2160 & 378 & 324 & 2862 & 7198 & 2099  & 1106 & 10403 & 22040 & 5358 & 4259 & 31657 & 9738 & 2757  & 683 & 13178 \\
\bottomrule
\end{tabular}
}
\vspace{-4pt}
\caption{ Breakdown of different datasets in 5 scenarios. ZT and ST denote Zero Target and Single Target, respectively.}
\vspace{-8pt}
\label{tab:datasets_breakdown}
\end{table*}

We conduct experiments on both ScanRefer~\cite{chen2020scanrefer} and \PROJECTNAME datasets and consider two setups: grounding objects with GT and predicted bounding boxes.

\xhdr{GT bounding boxes.}
For \OURMETHODCLIP and D3Net~\cite{chen2022d3net}, we input the complete scene and apply GT point masks for each object to get GT bounding boxes and masked features from the pre-trained detector (PointGroup for both D3Net and our \OURMETHODCLIP). %
For 3DVG-Trans~\cite{zhao20213dvg} and 3DJCG~\cite{cai20223djcg}, we use their original GT setting following the method proposed by ReferIt3D~\cite{achlioptas2020referit_3d}.
Single input objects are first extracted from the scene using the GT masks and PointNet++~\cite{qi2017pointnet++} is used for obtaining the object features.
For all methods, we disable the $L_{det}$ loss when using GT boxes.

\xhdr{Predicted bounding boxes.}
We use the original detector design of each method to predict bounding boxes and extract features.

\begin{table}
\centering
\resizebox{\linewidth}{!}
{
\begin{tabular}{@{}lrrr rrr@{}}
\toprule
& \multicolumn{3}{c}{Acc@0.5 on Val} & \multicolumn{3}{c}{Acc@0.5 on Test} \\
\cmidrule(lr){2-4}\cmidrule(lr){5-7}
 & Unique & Multiple & \textbf{All} & Unique & Multiple & \textbf{All} \\
\midrule
3DVG-Trans+~\cite{zhao20213dvg} & 62.0 & 30.3 & 36.4 & 57.9 & 31.0 & 37.0 \\
InstanceRefer~\cite{yuan2021instancerefer} & 66.8 & 24.8 & 32.9 & 66.7 & 26.9 & 35.8 \\
FFL-3DOG~\cite{feng2021free} & 67.9 & 25.7 & 34.0 & - & - & - \\
SAT~\cite{yang2021sat} & 50.8 & 25.2 & 30.1 & - & - & - \\
3D-SPS~\cite{luo20223d} & 66.7 & 29.8 & 37.0 & - & - & - \\
MVT~\cite{huang2022multi} & 66.5 & 25.3 & 33.3 & - & - & - \\
BUTD-DETR~\cite{jain2022bottom} & 66.3 & 35.1 & 39.8 & - & - & - \\
D3Net (G)~\cite{chen2022d3net} & 70.4 & 27.1 & 35.6 & 65.8 & 27.3 & 36.0 \\
D3Net* ~\cite{chen2022d3net} & 72.0 & 30.1 & 37.9 & 68.4 & 30.7 & 39.2 \\
3DJCG (G)~\cite{cai20223djcg} & 64.5 & 30.3 & 36.9 & - & - & - \\
3DJCG* ~\cite{cai20223djcg} & 64.3 & 30.8 & 37.3 & 60.6 & 31.2 & 37.8  \\
HAM~\cite{chen2022ham} & 67.9 & 34.0 & 40.6 & 63.7 & 33.2 & 40.1 \\
UniT3D (G)~\cite{chen2022unit3d} & 74.8 & 27.6 & 36.5 & - & - & - \\ 
UniT3D*~\cite{chen2022unit3d}  & 73.1 & 31.1 & 39.1 & - & - & - \\ 
\midrule
M3DRef-CLIP & \textbf{77.2} & \textbf{36.8} & \textbf{44.7} & \textbf{70.9} & \textbf{38.1} & \textbf{45.5} \\
\bottomrule
\end{tabular}
}
\vspace{-4pt}
\caption{
For unified models~\cite{chen2022d3net,cai20223djcg,chen2022unit3d}, we report both the grounding-only (G) performance as well as their best performance. We use * to indicate joint grounding and captioning models trained with extra data.}
\label{tab:scanrefer_comp}
\end{table}

\subsection{Implementation details}
We implement \OURMETHODCLIP using PyTorch Lightning\footnote{\href{https://www.pytorchlightning.ai/}{www.pytorchlightning.ai}} and  PyTorch3D~\cite{ravi2020pytorch3d}.
We train the model end-to-end on a single NVIDIA RTX A5000 with a batch size of 4, using the AdamW optimizer~\cite{loshchilov2017decoupled} with a learning rate of $5e^{-4}$.
We use a re-implementation of PointGroup\footnote{\href{https://github.com/3dlg-hcvc/minsu3d}{https://github.com/3dlg-hcvc/minsu3d}} with the Minkowski Engine~\cite{choy20194d}.
Following D3Net, we pretrain our PointGroup module on all ScanNet v2 training scans with the 18 ScanRefer categories.
We take point coordinates, point normals and per-point multi-view features $P \in \mathbb{R}^{N \times (3+3+128)}$ as the input.
For data augmentation, we randomly apply coordinate jitter, x-axis flipping and rotation around the z-axis.
We freeze a pre-trained CLIP with ViT-B/32~\cite{radford2021learning} and only train additional MLPs.
For encoding the text with CLIP, we follow CLIP and tokenize with a lower-cased BPE, and \texttt{[SOS]} and \texttt{[EOS]} tokens added (the output corresponding to \texttt{[EOS]} is used as the sentence representation).

\xhdr{Object renderer.}
For each object proposal, we render $3$ views horizontally spaced 120 degrees apart, at a distance of $1\text{m}$, with elevation angle of $45^\circ$.
We crop coordinates and colors from the input scene point cloud using predicted bounding boxes and set point radius to $2.5\text{cm}$ and image size to $224^{2}$.
We use CUDA to batch index scene point clouds and crop in parallel, and render the batches sparsely to avoid padding overhead.
Rendering is implemented with PyTorch3D and executed on the GPU. The computational overhead of our method with the online renderer ranges from $10-20\%$ 
(see \Cref{sec:supp:compute} for details).

\begin{table}
\centering
\resizebox{\linewidth}{!}
{
\begin{tabular}{@{}l rrrrr@{}}
\toprule
& Easy & Hard & View-Dep & View-Indep & \textbf{All} \\
\midrule
3DVG-Trans~\cite{zhao20213dvg} & 48.5 & 34.8 & 34.8 & 43.7 & 40.8 \\
InstanceRefer~\cite{yuan2021instancerefer} & 46.0 & 31.8 & 34.5 & 41.9 & 38.8 \\
FFL-3DOG~\cite{feng2021free} & 48.2 & 35.0 & 37.1 & 44.7 & 41.7 \\
TransRefer3D~\cite{he2021transrefer3d} & 48.5 & 36.0 & 36.5 & 44.9 & 42.1 \\
SAT~\cite{yang2021sat} & 56.3 & 42.4 & 46.9 & 50.4 & 49.2 \\
3D-SPS~\cite{luo20223d} & 58.1 & 45.1 & 48.0 & 53.2 & 51.5 \\
MVT~\cite{huang2022multi} & \textbf{61.3} & \textbf{49.1} & \textbf{54.3} & 55.4 & \textbf{55.4} \\
BUTD-DETR~\cite{jain2022bottom} & 60.7 & 48.4 & 46.0 & \textbf{58.0} & 54.6 \\
HAM~\cite{chen2022ham} & 54.3 & 41.9 & 41.5 & 51.4 & 48.2 \\
LanguageRefer~\cite{roh2022languagerefer} & 51.0 & 36.6 & 41.7 & 45.0 & 43.9 \\
LAR~\cite{bakr2022look} & 56.1 & 41.8 & 46.7 & 50.2 & 48.9 \\
\midrule
M3DRef-CLIP & 55.6 & 43.4 & 42.3 & 52.9 & 49.4  \\
\bottomrule
\end{tabular}
}
\vspace{-4pt}
\caption{Comparison of methods on Nr3D~\cite{achlioptas2020referit_3d} val set with GT boxes.
}
\vspace{-8pt}
\label{tab:scan_nr3d_comp}
\end{table}

\subsection{Results}

\subsubsection{Performance of \OURMETHODCLIP}

\begin{table}
\centering
\resizebox{\linewidth}{!}
{
\begin{tabular}{@{}llrrrrrrrrr@{}}
\toprule
& \multicolumn{3}{c}{Acc on ScanRefer} & \multicolumn{3}{c}{Acc on Multi3DRefer (ST)} \\
\cmidrule(lr){2-4} \cmidrule(lr){5-7}
Training Dataset & Unique & Multiple & \textbf{All} & Unique & Multiple & \textbf{All} \\
\midrule
ScanRefer~\cite{chen2020scanrefer} & 89.3 & 49.1 & 56.9 & 79.5 & 48.2 & 57.0 \\
Multi3DRefer (ST) & 88.5 & 46.9 & 55.0 & 86.3 & 52.9 & 62.3 \\
Multi3DRefer (ST) + ScanRefer & \textbf{90.8} & \textbf{51.0} & \textbf{58.7} & \textbf{88.0} & \textbf{56.7} & \textbf{65.5} \\

\bottomrule
\end{tabular}
}
\vspace{-4pt}
\caption{We compare results of training \OURMETHOD with GT boxes on different datasets' val set.
We only use the Single Target (ST) case in \PROJECTNAME dataset.}
\label{tab:scan_multi3d_comp}
\end{table}

\begin{table*}
\centering
\resizebox{\linewidth}{!}
{
\begin{tabular}{@{}lrrrrrr rrrrrr@{}}
\toprule
& \multicolumn{6}{c}{F1 (GT boxes)} & \multicolumn{6}{c}{F1@0.5 (Pred boxes)} \\
\cmidrule(lr){2-7}\cmidrule(lr){8-13}
 & ZT w/o D & ZT w/ D & ST w/o D & ST w/ D & MT & \textbf{All} & ZT w/o D & ZT w/ D & ST w/o D & ST w/ D & MT & \textbf{All} \\
\midrule
3DVG-Trans+~\cite{zhao20213dvg} & 45.3 & 14.3 & 58.9 & 35.2 & 54.2 & 44.1 & 87.1 & 45.8 & 27.5 & 16.7 & 26.5 & 25.5  \\
D3Net (Grounding)~\cite{chen2022d3net} & 71.6 & 20.4  & 78.2 & 44.4 & 61.6 & 55.5 & 81.6 & 32.5 & 38.6 & 23.3 & 35.0 & 32.2 \\
3DJCG (Grounding)~\cite{cai20223djcg} & 47.9 & 16.4 & 59.1 & 35.5 & 54.2 & 44.6 & \textbf{94.1} & \textbf{66.9} & 26.0 & 16.7 & 26.2 & 26.6 \\
\midrule
M3DRef-CLIP & \textbf{74.2} & \textbf{29.4} & \textbf{84.1} & \textbf{52.3} & \textbf{67.2} & \textbf{62.3} & 81.8 & 39.4  & \textbf{47.8} & \textbf{30.6} & \textbf{37.9} & \textbf{38.4} \\
\bottomrule
\end{tabular}
}
\vspace{-4pt}
\caption{Comparison of different methods on \PROJECTNAME. Our \OURMETHODCLIP outperforms prior work on most metrics.}
\label{tab:multi3drefer_comp}
\vspace{-8pt}
\end{table*}

We first validate the performance of \OURMETHODCLIP on ScanRefer (see \Cref{tab:scanrefer_comp}) and Nr3D~\cite{achlioptas2020referit_3d} (see \Cref{tab:scan_nr3d_comp}) datasets and compare it against recent models.
On ScanRefer, our method outperforms all prior works including those joint models leveraging extra input data by a large margin on both val set and test set (online benchmarking).
In our ablation study, we find that the use of the CLIP text encoder is a key factor to the strong performance of our model on ScanRefer.
Another key factor is the use of a strong 3D object instance segmentation network (PointGroup) as our object detector.
For instance, the PointGroup-based methods~\cite{chen2022d3net,chen2022unit3d} readily outperform VoteNet-based methods~\cite{zhao20213dvg,cai20223djcg} on the unique subset of ScanRefer.

For Nr3D, we achieve comparable but less competitive results because of our weaker ground-truth box encoder.
Note that SAT~\cite{yang2021sat} and MVT~\cite{huang2022multi} also leverage 2D images but render them offline.
Overall, using additional 2D image information is helpful for two-stage methods.

\begin{table}
\centering
\resizebox{\linewidth}{!}
{
\begin{tabular}{@{}lrrr rrr@{}}
\toprule
& \multicolumn{3}{c}{Acc (GT boxes)} & \multicolumn{3}{c}{Acc@0.5 (Pred boxes)} \\
\cmidrule(lr){2-4}\cmidrule(lr){5-7}
Eval Dataset & Unique & Multiple & \textbf{All} & Unique & Multiple & \textbf{All} \\
\midrule
\PROJECTNAME (ST) & 86.9 & 51.5 & 61.5 & 67.3 & 40.1 & 47.7 \\
ScanRefer~\cite{chen2020scanrefer} & 88.0 & 46.1 & 54.3 & 73.5 & 34.3 & 41.9 \\
\bottomrule
\end{tabular}
}
\vspace{-4pt}
\caption{\OURMETHODCLIP evaluated on \PROJECTNAME (ST) and ScanRefer with both GT and predicted bounding boxes.}
\label{tab:eval_multi_scanrefer}
\end{table}

\subsubsection{\PROJECTNAME}

We split the \PROJECTNAME data into train/val/test by scene following the ScanRefer split, resulting in a rough split ratio of 7:2:1 for the scene-description pairs.
See \Cref{tab:datasets_breakdown} for statistics, including the number of descriptions with zero, single, or multiple targets.

\xhdr{Usefulness of \PROJECTNAME dataset.}
To study the usefulness of the \PROJECTNAME dataset, we compare the performance of training \OURMETHODCLIP with the original ScanRefer data and with our \PROJECTNAME data for the Single Target (ST) case.
We evaluate on both ScanRefer and \PROJECTNAME (ST), using GT bounding boxes.
\Cref{tab:scan_multi3d_comp} shows that our reworded data can improve performance on ScanRefer.
Prior work has also shown that incorporating additional labelled data can help, but they typically used either a captioning model~\cite{chen2022d3net,chen2022unit3d}, mixed additional annotated data~\cite{yang2021sat}, or used dense annotations~\cite{abdelreheem2022scanents3d}.
We show that simple rewordings (without accessing the images) also help.

We also evaluate \OURMETHODCLIP trained on \PROJECTNAME using ScanRefer's task setting (\Cref{tab:eval_multi_scanrefer}) of predicted objects.
We observe that the model trained on \PROJECTNAME data and task achieves similar performance to the model trained on ScanRefer data and task, which illustrates the generalization of \PROJECTNAME.

\xhdr{Evaluation on \PROJECTNAME.}
We compare four models on the \PROJECTNAME dataset using both ground-truth and predicted boxes (\Cref{tab:multi3drefer_comp}).
In our experiments, we focus on two-stage methods that perform well on ScanRefer.
We adapt the code of 3DVG-Trans, 3DJCG and D3Net to our task.
For 3DVG-Trans, we use the enhanced version 3DVG-Trans+ provided by the authors.\footnote{\href{https://github.com/zlccccc/3DVG-Transformer}{github.com/zlccccc/3DVG-Transformer}}
We only train and evaluate the grounding model of 3DJCG and D3Net.

\begin{figure}
\includegraphics[width=\linewidth]{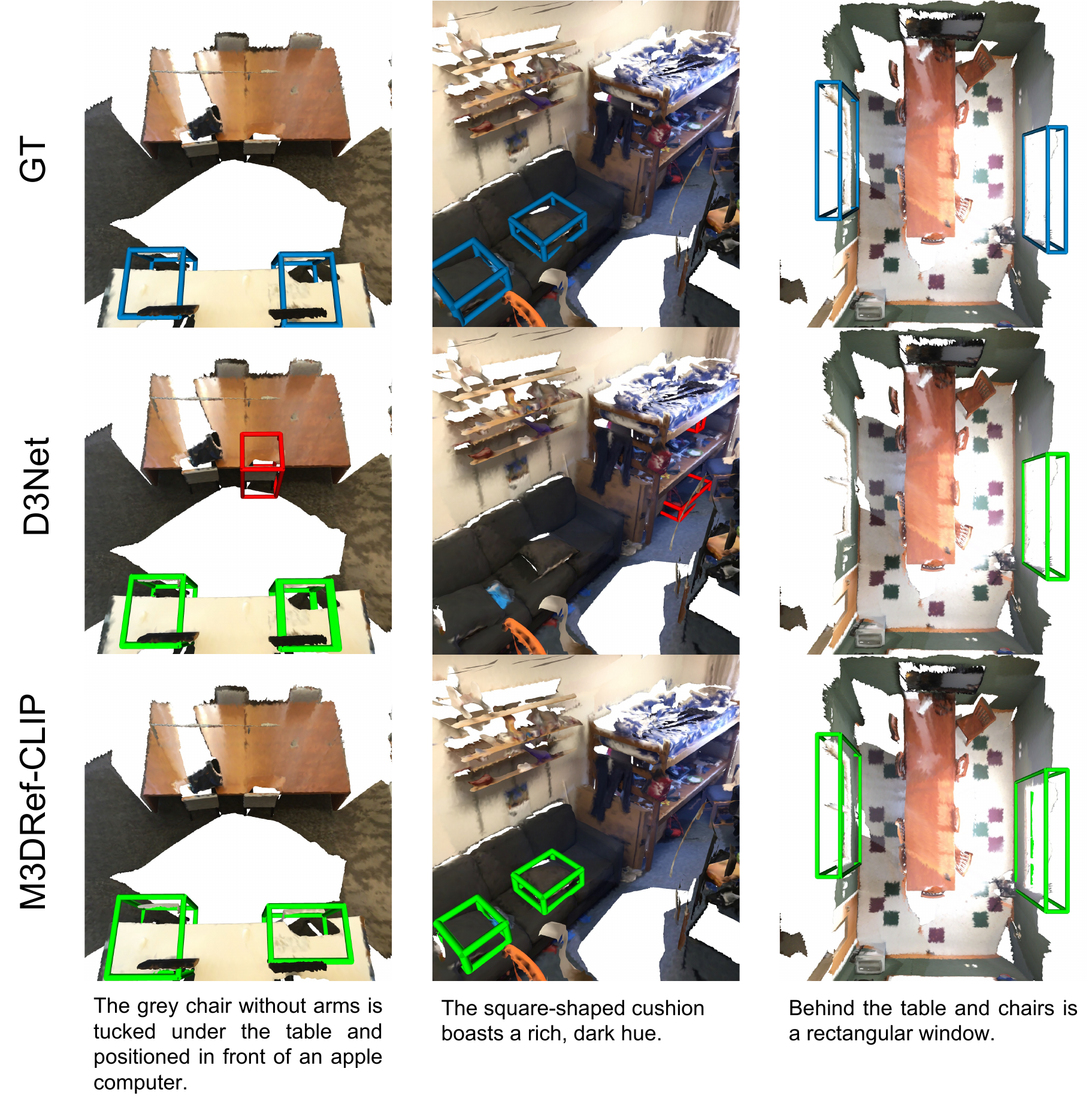}
\vspace{-20pt}
\caption{Qualitative results of D3Net~\cite{chen2022d3net} versus \OURMETHODCLIP on \PROJECTNAME using predicted boxes.
Blue boxes indicate GT, green boxes are true positives with IoU threshold \predth $>0.5$.
Red boxes are false positives.}
\label{fig:qualitative}
\vspace{-10pt}
\end{figure}

To analyze the performance of the models, we break down the zero (ZT) and single-target (ST) case to without and with distractors (D) of the same class.
Overall, having distractors is more challenging.
We note that \OURMETHODCLIP outperforms the other methods on \PROJECTNAME, and that 3DJCG is better at handling the ZT case with predicted boxes.
\Cref{fig:qualitative} shows qualitative results (see \Cref{sec:supp:qual_results} for more results).

\begin{table*}[t]
\centering
{
\begin{tabular}{@{}lllrrr rrrrrr@{}}
\toprule
& & & \multicolumn{3}{c}{Acc (ScanRefer)} & \multicolumn{6}{c}{F1 (Multi3DRefer)} \\
\cmidrule(lr){4-6} \cmidrule(lr){7-12}
& Text & Vision & Uniq & Mult & \textbf{All} & ZT w/o D & ZT w/ D & ST w/o D & ST w/ D & MT & \textbf{All} \\

\midrule %
\multirow{3}{*}{\rotatebox[origin=c]{90}{GT}}
& GRU & 3D & 88.8 & 43.5 & 52.3 & 70.1 & 27.3 & 81.6 & 49.3 & 63.3 & 59.1 \\
& CLIP & 3D & 87.9 & 48.3 & 56.0 & 71.8 & \textbf{29.9} & 83.5 & 51.6 & 65.7 & 61.3 \\
& CLIP & CLIP & 78.9 & 42.1 & 49.3 & 58.3 & 27.8 & 74.1 & 44.7 & 61.0 & 54.4 \\
& CLIP & 3D+CLIP & \textbf{89.3} & \textbf{49.1} & \textbf{56.9}  & \textbf{74.2} & 29.4 & \textbf{84.1} & \textbf{52.3} & \textbf{67.2} & \textbf{62.3} \\

\midrule
\multirow{3}{*}{\rotatebox[origin=c]{90}{Pred}}
& GRU & 3D & 72.2 & 32.8 & 40.4 & 78.8 & 42.1 & \textbf{49.4} & 28.5 & 37.0 & 37.4 \\
& CLIP & 3D & 75.2 & 34.7 & 42.6 & 77.1 & 34.1 & 48.8 & 30.0 & \textbf{39.2} & 38.2 \\
& CLIP & CLIP & 71.2 & 31.2 & 39.0 & 64.4 & 36.0 & 42.9 & 25.7 & 33.7 & 33.1 \\
& CLIP & 3D+CLIP & \textbf{77.2} & \textbf{36.8} & \textbf{44.7} & \textbf{81.8} & \textbf{39.4} & 47.8 & \textbf{30.6} & 37.9 & \textbf{38.4} \\

\bottomrule
\end{tabular}
}
\vspace{-4pt}
\caption{Ablations in training \OURMETHOD using different feature embeddings and ground-truth boxes (top rows) and predicted boxes (bottom rows). The combination of CLIP~\cite{radford2021learning} and 3D features achieves the best performance.  We use PointGroup~\cite{jiang2020pointgroup} for our 3D object detector and feature extractor.}
\label{tab:ablation_clip}
\vspace{-8pt}
\end{table*}

\begin{table*}
\centering
\resizebox{\linewidth}{!}
{
\begin{tabular}{@{}lrrr rrrrrr rrrrrr@{}}
\toprule
& \multicolumn{3}{c}{ScanRefer} & \multicolumn{5}{c}{Nr3D} & \multicolumn{6}{c}{Multi3DRefer} \\
\cmidrule(lr){2-4} \cmidrule(lr){5-9} \cmidrule(lr){10-15}
 & Unique & Multiple & \textbf{All} & Easy & Hard & View-Dep & View-Indep & \textbf{All} & ZT w/o D & ZT w/ D & ST w/o D & ST w/ D & MT & \textbf{All} \\
\midrule
w/ contrastive & 89.3 & \textbf{49.1} & \textbf{56.9} & \textbf{55.6} & \textbf{43.4} & \textbf{42.3} & \textbf{52.9} & \textbf{49.4} & \textbf{74.2} & \textbf{29.4} & \textbf{84.1} & \textbf{52.3} & \textbf{67.2} & \textbf{62.3} \\
w/o contrastive & \textbf{89.6} & 47.2 & 55.4 & 50.9 & 35.3 & 37.6 & 45.6 & 42.9 & 67.4 & 23.8 & 83.4 & 52.0 & 66.5 & 61.1 \\
\bottomrule
\end{tabular}
}
\vspace{-4pt}
\caption{Ablation of \OURMETHODCLIP with and without contrastive loss using GT boxes.}
\label{tab:ablation_contrastive}
\vspace{-8pt}
\end{table*}

\subsubsection{Ablation studies}
\xhdr{Does CLIP help?}
We experiment with just using pure CLIP for both text and image encoding vs combining CLIP and 3D features (see \Cref{tab:ablation_clip}).
We also report results using a GRU~\cite{cho2014learning}-based text encoder with GloVE~\cite{pennington2014glove} embedding, as used in prior work~\cite{chen2020scanrefer,zhao20213dvg,chen2022d3net}.
Although our GRU and 3D variants are similar at a high level to the grounding model of D3Net (which also used GRU and PointGroup with a transformer-based fusion module), we outperform the D3Net grounding module.
\Cref{tab:ablation_clip} shows that using the CLIP text encoder improves performance, and combining CLIP and 3D features yields the best performance.
The CLIP image encoder by itself underperforms the 3D features, showing the usefulness of 3D information.

\xhdr{Does contrastive learning help?}
\Cref{tab:ablation_contrastive} reports the performance of \OURMETHOD with and without contrastive learning on ScanRefer, Nr3D and \PROJECTNAME.
We observe the benefit of the contrastive loss for all tasks, especially Nr3D.

\subsection{Analysis of \OURMETHODCLIP}

To better study the \PROJECTNAME data and understand how \OURMETHODCLIP helps address the task, we break down evaluation of \PROJECTNAME based on attributes provided in the description: spatial, color, texture, and shape information.
For this analysis, these four splits are mutually exclusive, i.e. we only keep descriptions that describe exactly 1 attribute from the four and discard others.
We compare using \OURMETHOD with GRU with 3D PointGroup features vs our full \OURMETHODCLIP model.
We use the GT boxes setting and report F1 scores in \Cref{tab:attr_comp_suppl}.
We observe that adding features from CLIP helps identify all these attributes, with the mix of 3D+CLIP giving the best performance.
We found that descriptions with spatial terms are more challenging than descriptions with texture or shape with the addition of CLIP image features helping the most with color and shape descriptions.

\begin{table}
\centering
{
\begin{tabular}{@{}llrrrr@{}}
\toprule
Text & Vision & Spatial & Color & Texture & Shape\\
\midrule

GRU & 3D & 55.8 & 67.7 & 76.4 & 77.8 \\
CLIP & 3D & 58.6 & 68.9 & 79.2 & 78.5 \\
CLIP & CLIP & 50.0 & 64.4 & 73.9 & 74.1 \\
CLIP & 3D+CLIP & \textbf{58.8} & \textbf{71.9} & \textbf{79.4} & \textbf{82.5} \\
\bottomrule
\end{tabular}
}
\vspace{-4pt}
\caption{Breakdown of \OURMETHOD performance on descriptions with different attributes. We report F1 scores with GT boxes on \PROJECTNAME val set.}
\label{tab:attr_comp_suppl}
\end{table}

\section{Conclusion}

We present \PROJECTNAME, a more realistic task of grounding a flexible number of objects in real-world 3D scenes using natural language descriptions.
We designed a simple and efficient data generation pipeline to create data with less human effort, by leveraging existing language data and ChatGPT.
With this pipeline, we created a more diverse dataset consisting of more natural descriptions of varying granularity.
We also explored an end-to-end baseline method for solving the new task, which enables the online rendering of proposal objects to generate 2D cues, it also demonstrated the usefulness of CLIP~\cite{radford2021learning} and the multi-modal contrastive loss.
We believe \PROJECTNAME will bring more challenges and practical value in the direction of bridging 3D vision and language, especially for robotics and embodied AI tasks.

\xhdr{Future Work.}
Our current design relies on features from the 3D object detector to capture the global context, and the 2D image encoder to capture per-object attributes. Using positional encoding could improve the ability of the model to handle spatial relations. Investigating whether positional encoding improves the model, and what kind of positional encoding works best is a great avenue for future work.
\section*{Acknowledgements}
This work was funded in part by a Canada CIFAR AI Chair and NSERC Discovery Grant, and enabled in part by support from 
\href{https://alliancecan.ca/en}{the Digital Research Alliance of Canada}.
We thank Zhenyu (Dave) Chen for providing us with the ambiguous descriptions initially collected for ScanRefer, and Manolis Savva for proofreading and editing suggestions.
We also thank Archita Srivastava, Cody Ning, and Austin T. Wang for helping to verify the Multi3DRefer dataset.

{\small
\bibliographystyle{plainnat}
\setlength{\bibsep}{0pt}
\bibliography{main}
}

\clearpage
\newpage 
\section*{Appendix}
\appendix
\vspace{-4pt}

\begin{figure}
  \includegraphics[width=\linewidth]{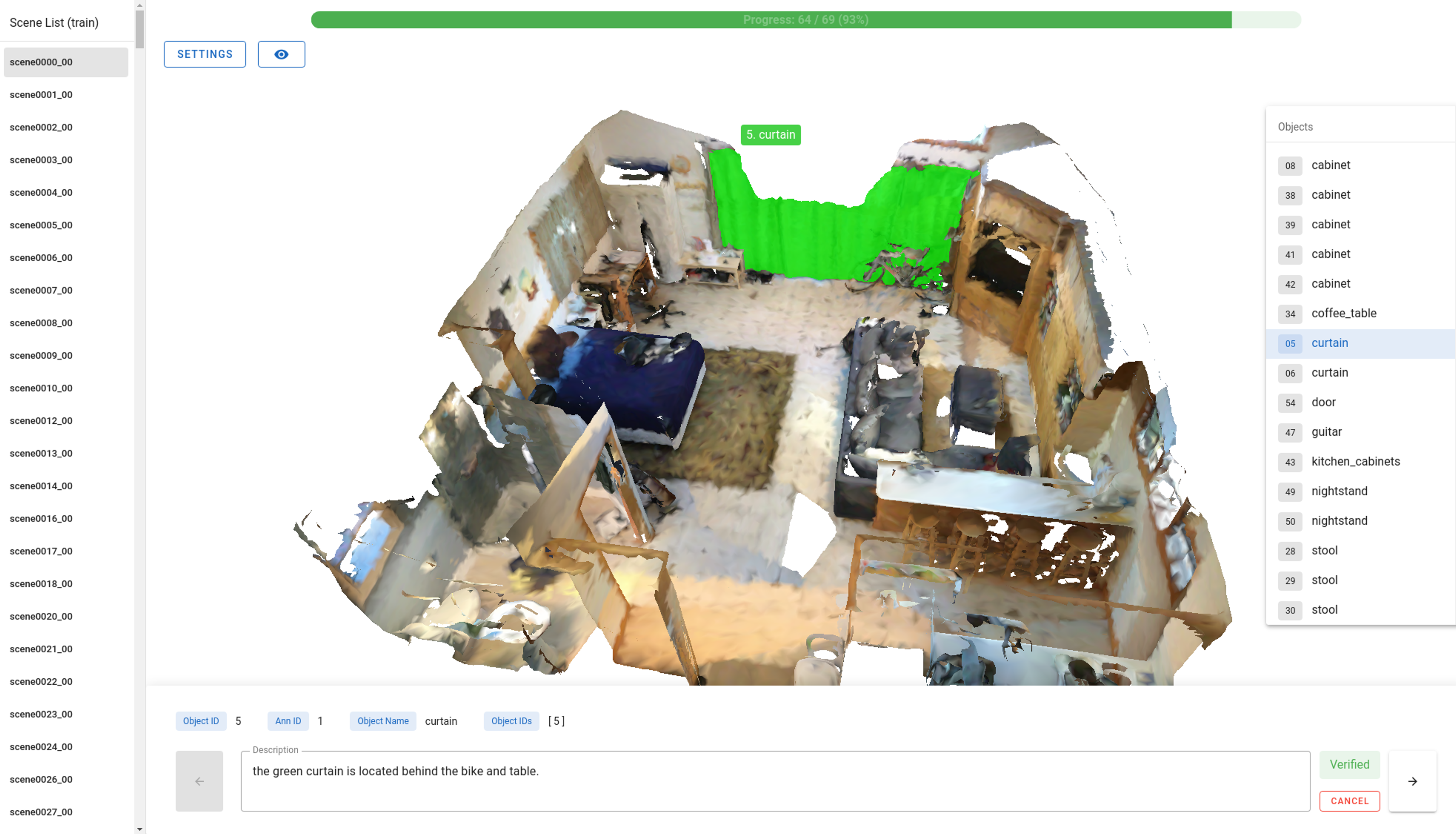}
  \vspace{-20pt}
  \caption{Using the \PROJECTNAME verification web interface,  verifiers check whether that the description matches the selected objects, and modify the list of selected objects or the description if needed.  The interface allows the verifiers to look at the scene from different viewpoints, and select objects by clicking.}
  \label{fig:web_screenshot}
  \vspace{-14pt}
\end{figure}

In this appendix, we provide more details about the web interface for verification (\Cref{sec:supp:web}), statistics (\Cref{sec:supp:data}) and qualitative examples from our dataset (\Cref{fig:supp:dataset_qual}).  We also discuss the computational efficiency of our online renderer (\Cref{sec:supp:compute}), analysis of matching strategies and thresholds on the Multi3DRefer task (\Cref{sec:supp:analysis}), and provide additional qualitative examples of our \OURMETHODCLIP (\Cref{sec:supp:qual_results}).

\section{Web interface for verification}
\label{sec:supp:web}
\begin{figure}
\centering
\setkeys{Gin}{width=\linewidth}
\begin{tabularx}{\linewidth}{@{}Y Y Y@{}}
\toprule
\includegraphics[trim=0 0 0 0,clip]{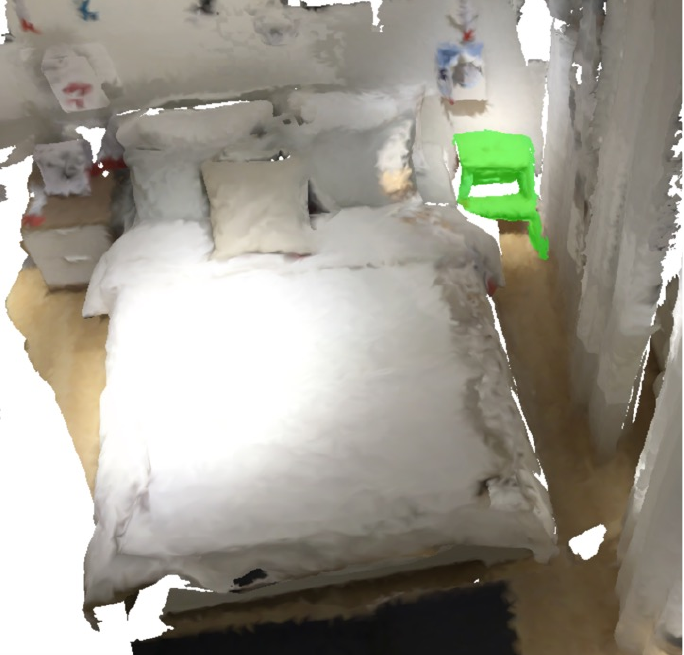} &
\includegraphics[trim=0 0 0 0,clip]{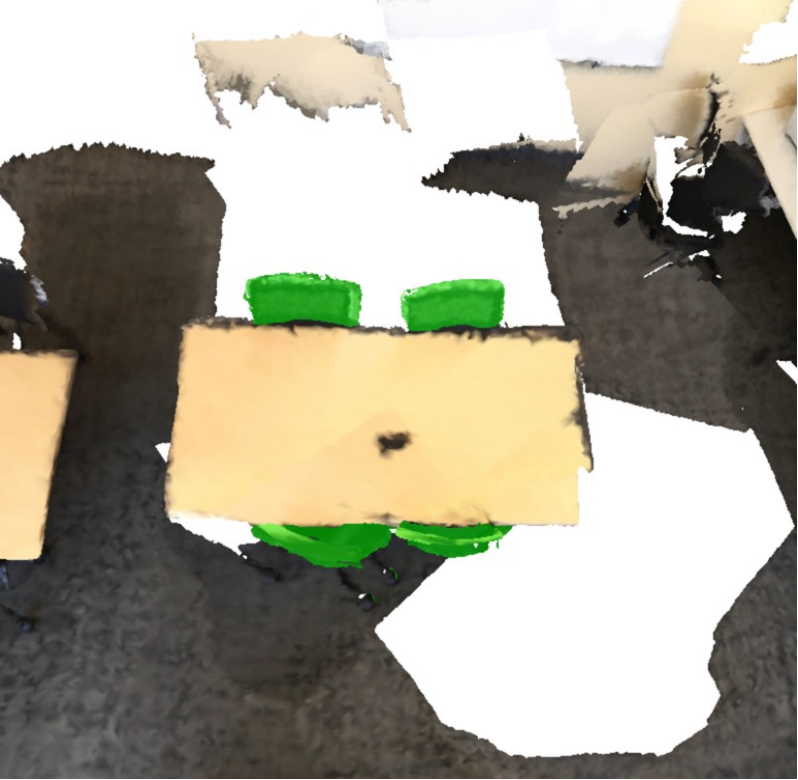} &
\includegraphics[trim=0 30 0 20,clip]{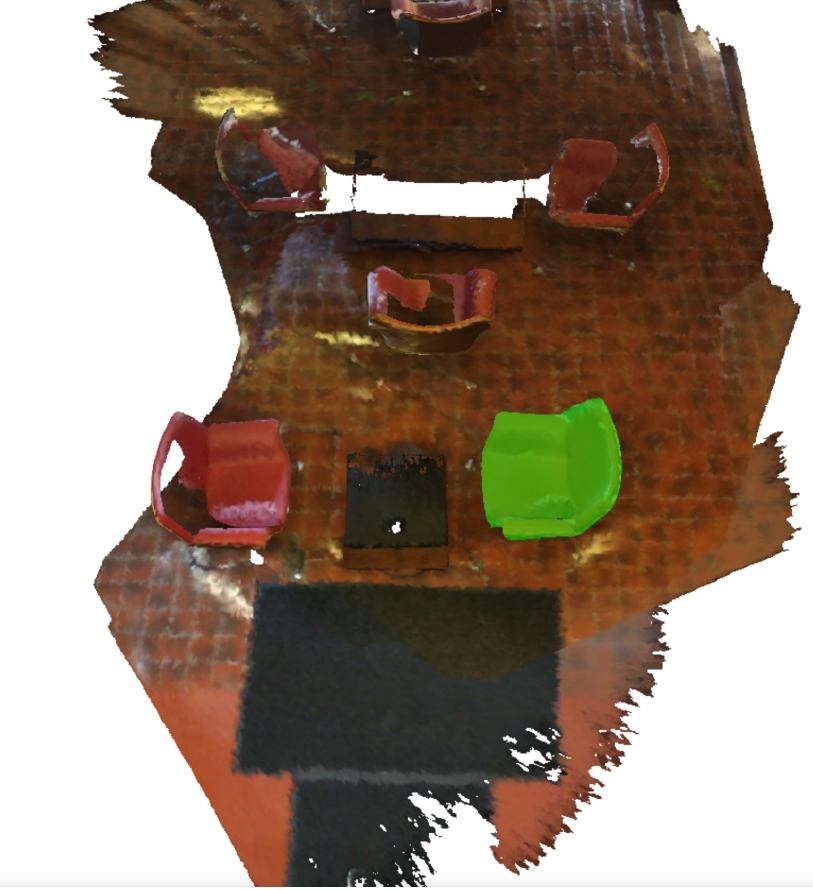} \\
\scriptsize{(a) The target stool is to the right of the bed.} &
\scriptsize{(b) The target chair is to the right of the table.} &
\scriptsize{(c) The chair is located to the right of you, as you stand on the black rug and face the coffee table.} 
\\
\bottomrule
\end{tabularx}
\vspace{-8pt}
\caption{ 
Examples of descriptions with spatial relation shown to annotators with target objects when (a) a single object matches, (b) multiple objects can match depending on where the viewer would be, and (c) a case where the viewpoint is specified in the description.
}
\label{fig:supp:web_examples}
\vspace{-6pt}
\end{figure}

\vspace{-6pt}
We implement a web-based data verification application using Three.js\footnote{\href{https://threejs.org/}{https://threejs.org/}}, Vue.js\footnote{\href{https://vuejs.org/}{https://vuejs.org/}} and FastAPI\footnote{\href{https://fastapi.tiangolo.com/}{https://fastapi.tiangolo.com/}}, to allow human verifiers to verify and correct the generated data.
See \Cref{fig:web_screenshot} for a screenshot of our web interface.
Verifiers are shown a generated description together with an interactive 3D mesh of a scene, where the selected objects are highlighted in green.
Verifiers are asked to check whether the description matches the identified target objects (in green).  If the description does not match, verifiers are asked to either: 1) change the target object list (by clicking on objects in the scene to toggle selection); or 2) modify the description if necessary.
Once the description clearly matches the selected objects, the `Verify' button is clicked to indicate that the pair has been manually verified.
Verifiers are instructed to consider whether a viewpoint is specified in the description or not.
If a specific viewpoint is given, then the viewpoint should be used to identify the specific objects being described.
If no viewpoint is given, then annotators are instructed to imagine different potential viewpoints from which they can stand and all objects that can match the given description.
See \Cref{fig:supp:web_examples} for examples shown to annotators. 

\begin{figure*}
\centering
\setkeys{Gin}{width=\linewidth}

\begin{tabularx}{\linewidth}{@{} 
  p{\dimexpr.40\linewidth-2\tabcolsep-1.3333\arrayrulewidth}%
  p{\dimexpr.60\linewidth-2\tabcolsep-1.3333\arrayrulewidth}%
@{}}
\raisebox{1.5\normalbaselineskip}[0pt][0pt]{
\rotatebox{90}{Zero Target}}
\includegraphics[trim=30 90 50 52,clip]{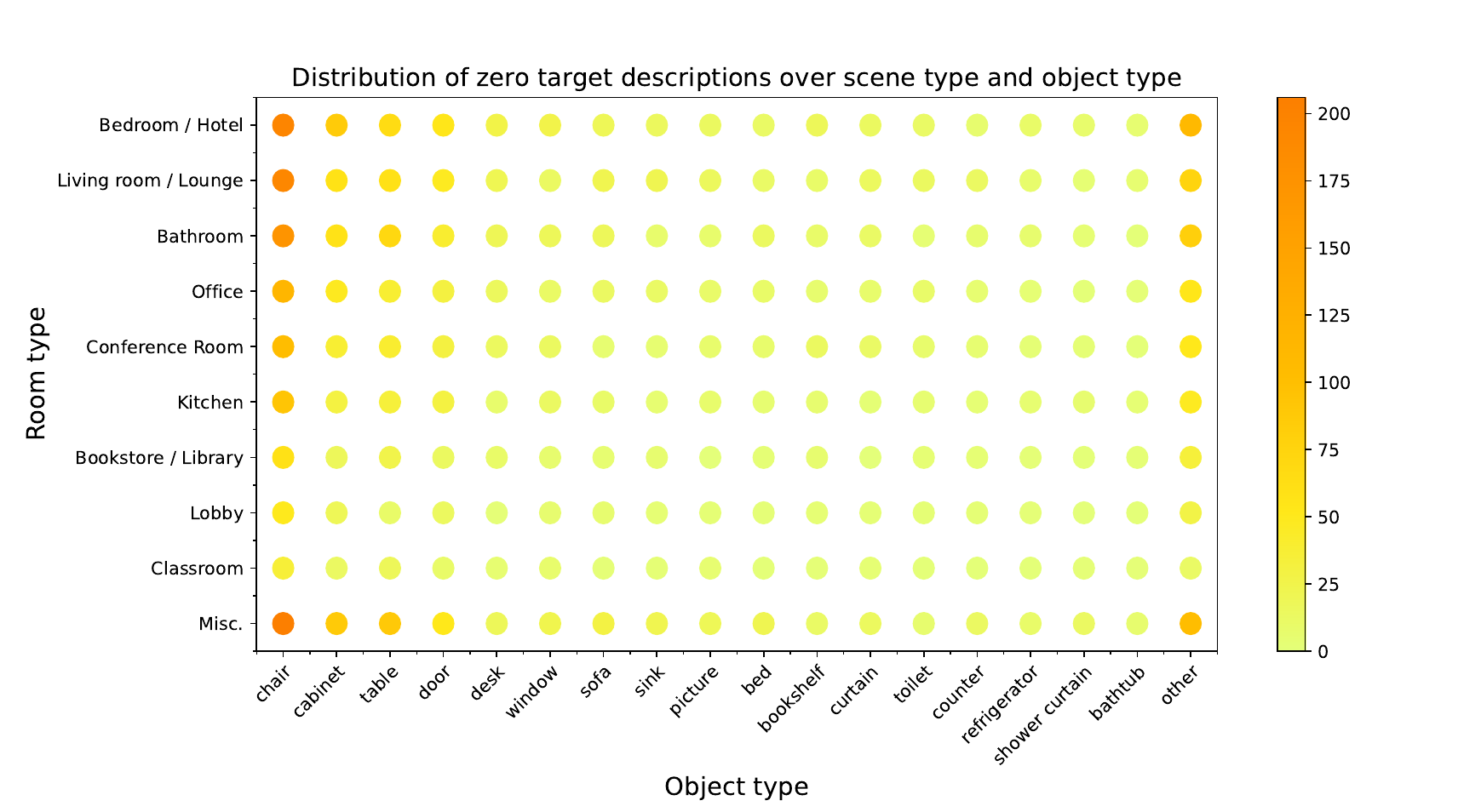} &  
\multirow{2}{\hsize}{
\raisebox{0.0\height}[0pt][0pt]{\makecell[c]{Multi Target\\ \includegraphics[trim=30 25 50 52,clip]{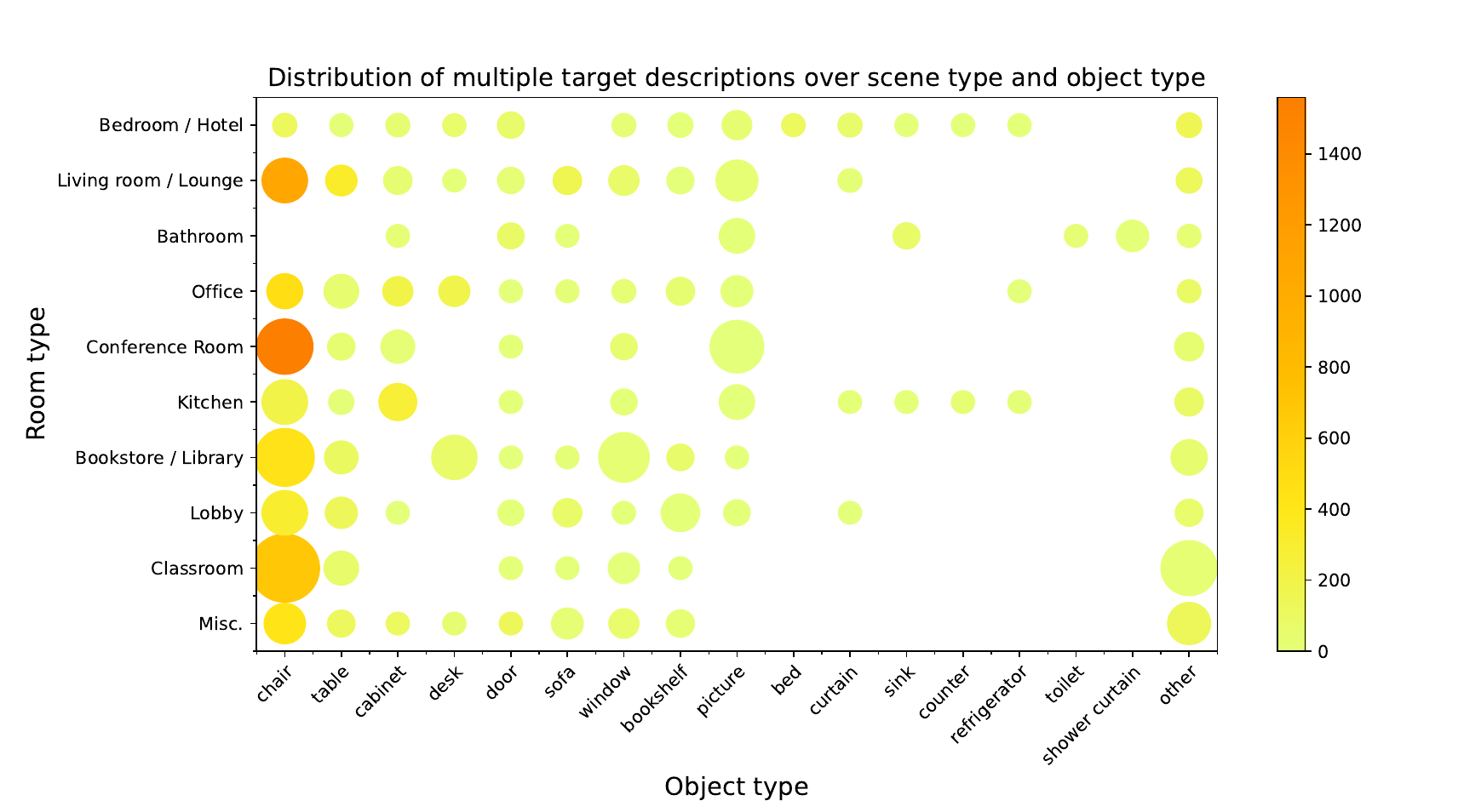}}}} \\ 
\raisebox{2.5\normalbaselineskip}[0pt][0pt]{
\rotatebox{90}{Single Target}}
\includegraphics[trim=30 25 50 52,clip]{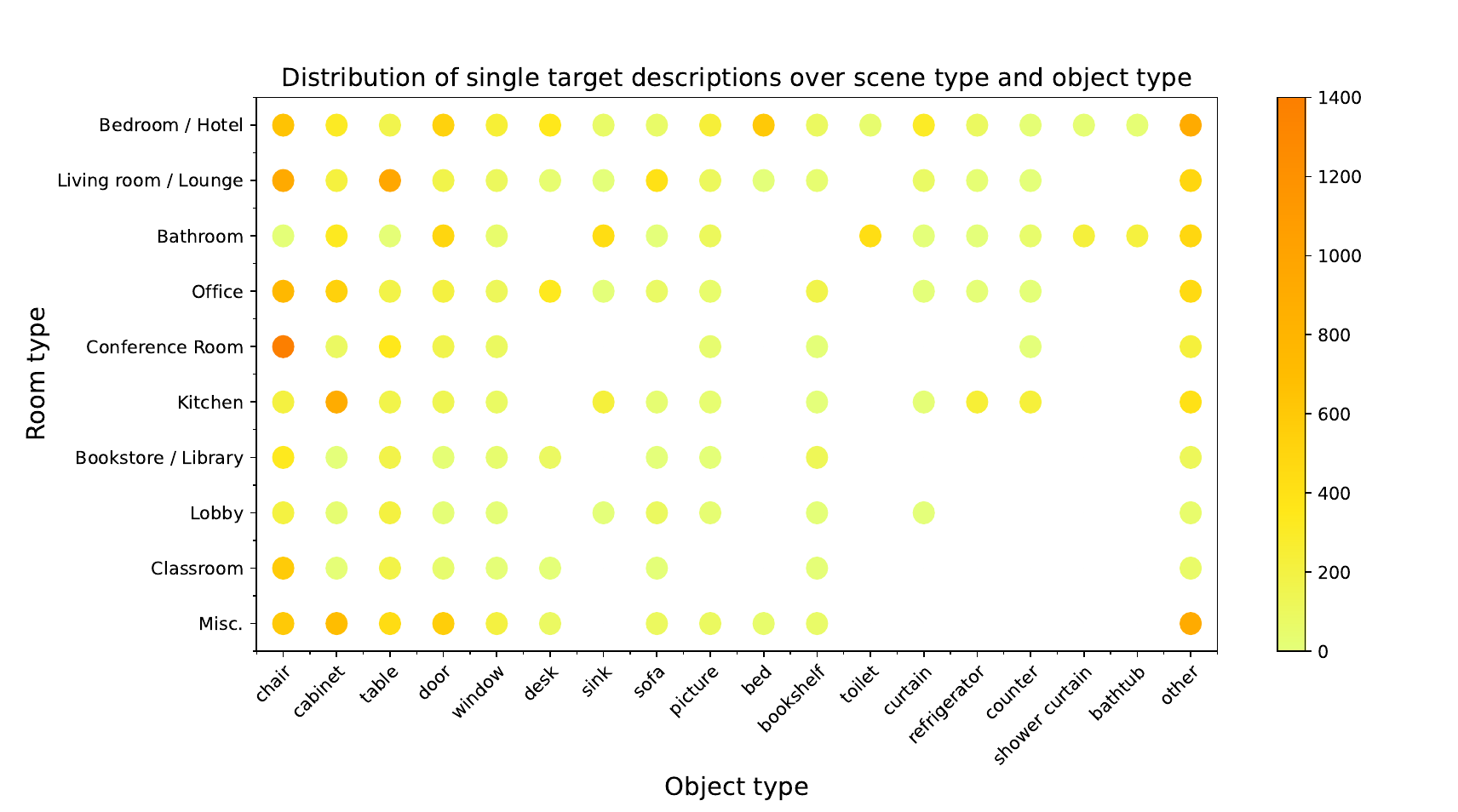} & \\
\end{tabularx}
\vspace{-12pt}
\caption{ 
Distribution of number of descriptions (color) and average number of target objects per description (circle size) by scene type and object type.  For Zero Target, the descriptions are evenly distributed across the different scene and object types. For Single Target descriptions, the distribution of descriptions over scene-object types reflects the distribution of real-world objects in scenes (e.g. shower curtains and bathtubs are found in bathrooms and hotel rooms).  For Multiple Targets descriptions, the size of the circle indicates the average number of target objects per description, ranging from 2 (for the smallest circle) to 5.8 (for the largest circle of chairs in classrooms), with a maximum of 32 targets for a single description.  The most common Multiple Targets descriptions are about chairs in classrooms, conference rooms, and libraries, while the least common are for objects where there is typically just one in a scene, but sometimes there are still multiple instances (e.g. sinks and refrigerators in kitchens).  Note that bathtubs are omitted from the Multiple Targets case, as in this dataset, there are no scenes with more than one bathtub.
}
\label{fig:supp:stats_scene_object}
\vspace{-16pt}
\end{figure*}

In total, verifiers checked 64513 description-scene pairs.  Of these, we discard 2587 samples (542 from Zero Target and 2045 from Multiple Targets) to limit the number of zero-target descriptions per scene to 21
and the number of overly similar descriptions for complex scenes.  During verification, 11804 descriptions were modified by verifiers.  %
Most of the modifications were minor changes such as changing `left' to `right', or adding more constraints (e.g. changing `this is a chair' to `this is a chair facing the wall').
The verification check took about 9 seconds per zero target description and 16 seconds per Single Target / Multiple Targets description.

\section{Statistics and examples of \PROJECTNAME}
\label{sec:supp:data}
\begin{figure*}
  \centering
  \includegraphics[width=\linewidth]{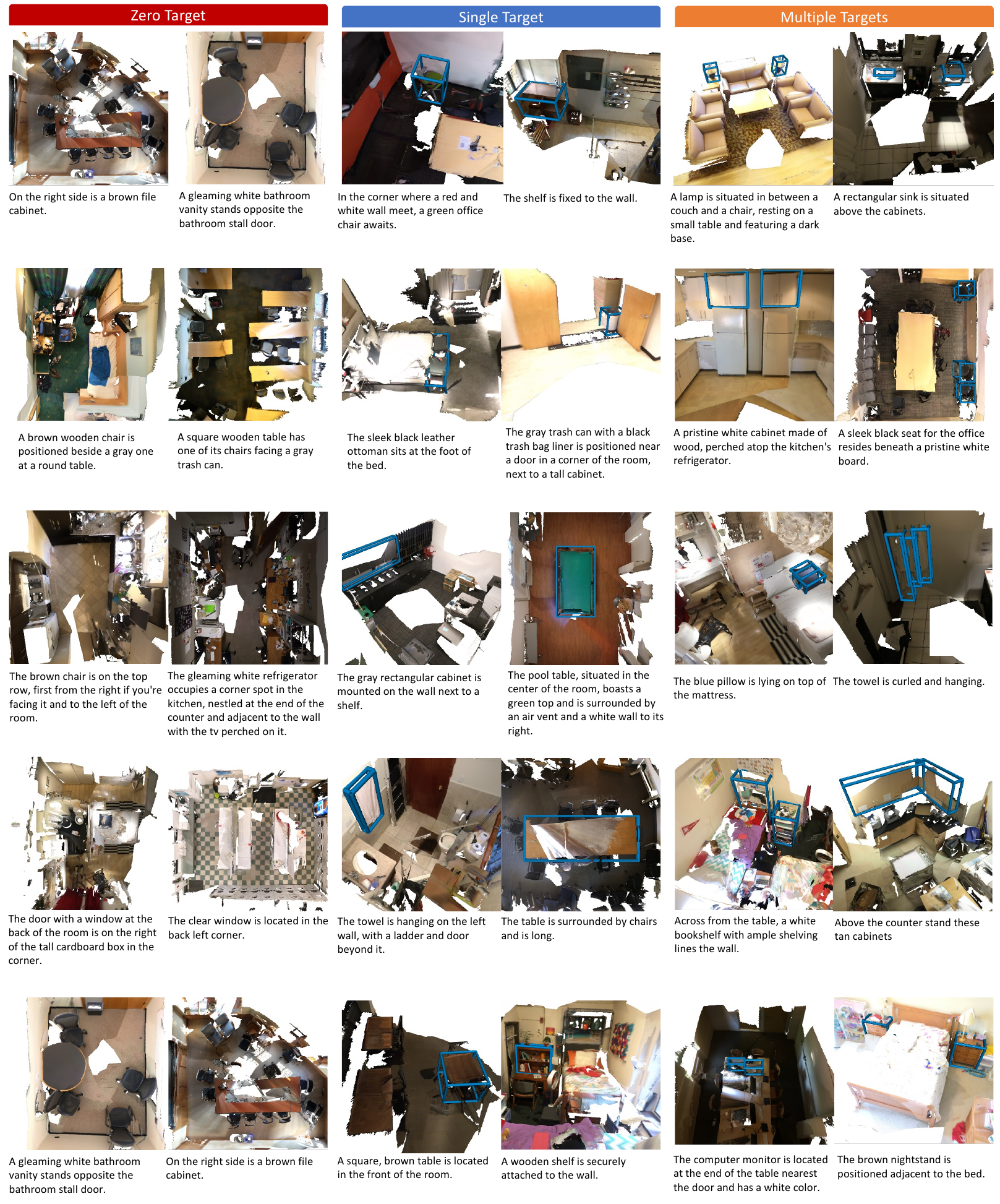}
  \vspace{-18pt}
  \caption{Examples of scene-description pairs with Zero Target, Single Target, and Multiple Targets from our \PROJECTNAME dataset. Blue boxes indicate GT.}
  \label{fig:supp:dataset_qual}
\end{figure*}

\vspace{-6pt}
In \Cref{tab:stat_multi3drefer_suppl}, we show the overall number of descriptions, scenes, and objects across the train/val/test split in the \PROJECTNAME dataset. We visualize the distribution of the number of descriptions and the average of target objects per description broken down by scene type and object type in \Cref{fig:supp:stats_scene_object}.  We see that the Single Target descriptions reflect the distributions of objects in the real world, while the Zero Target descriptions are more evenly distributed.  For the Multiple Targets descriptions, chairs are most common with the number of targets ranging from 2 to 32.
\Cref{fig:supp:dataset_qual} show specific examples of the \PROJECTNAME dataset with descriptions matching zero, single, or multiple targets.

\begin{table}
\centering
\resizebox{\linewidth}{!}
{
\begin{tabular}{@{}lrrrr@{}}
\toprule
 & Train & Val & Test & \textbf{Total} \\
\midrule
\#descriptions & 43,838 & 11,120 & 6,968 & 61,926 \\
\#scenes & 562 & 141 & 97 & 800 \\
\#objects & 8,346 & 2,161 & 1,102 & 11,609 \\
avg. \#objects / scene & 14.9 & 15.3 & 11.4 & 14.5 \\
avg. \#descriptions / scene & 78.0 & 78.9 & 71.8 & 77.4 \\
avg. \#descriptions / object & 5.3 & 5.1 & 6.3 & 5.3 \\
\bottomrule
\end{tabular}
}
\vspace{-8pt}
\caption{\PROJECTNAME statistics on train/val/test splits.}
\label{tab:stat_multi3drefer_suppl}
\vspace{-16pt}
\end{table}

\begin{table}
\centering
{
\begin{tabular}{lrrrr}
\toprule
& \multicolumn{2}{c}{Training} & \multicolumn{2}{c}{Inference} \\
\cmidrule(l{2pt}r{0pt}){2-3} 
\cmidrule(l{2pt}r{0pt}){4-5} 
& Mem & Time & Mem & Time \\
\midrule
D3Net (Grounding) & 14.7G & 41.1m & 15.2G & 10.1m \\
\OURMETHODCLIP & 15.2G & 55.5m & 11.3G & 12.5m \\
\bottomrule
\end{tabular}
}
\vspace{-6pt}
\caption{Comparison of GPU memory usage and running time between D3Net~\cite{chen2022d3net} and \OURMETHODCLIP.}
\label{tab:compute_efficiency}
\vspace{-14pt}
\end{table}

\section{Computational efficiency}
\label{sec:supp:compute}
\vspace{-6pt}
We compare training and inference time and GPU memory usage with the D3Net~\cite{chen2022d3net} grounding module (which also uses PointGroup~\cite{jiang2020pointgroup} as the detector) using \textit{torch.cuda.max\_memory\_reserved} (\Cref{tab:compute_efficiency}).
We use the same input with batch size 4 for 60 epochs until convergence and report GPU memory and time per epoch for the same machine with an NVIDIA RTX A5000 GPU.
The memory and computation overhead is only 10-20\%, including all rendering.

\begin{figure*}

\begin{subfigure}{\linewidth}

\includegraphics[width=\linewidth]{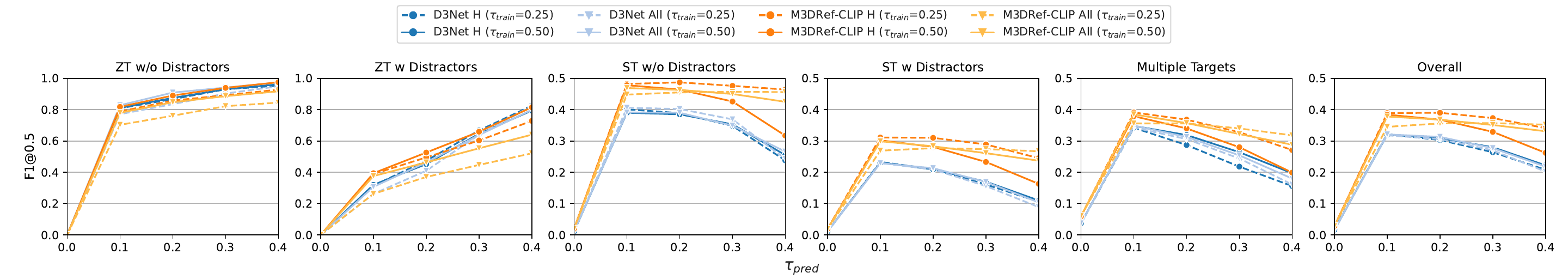}
\end{subfigure}
\vspace{-18pt}
\caption{F1@0.50 on \PROJECTNAME for the two methods with different matching strategies during training (All, Hungarian) and different values of \predth (x-axis), \ioutrainth (solid=0.5, dashed=0.25). As we increase the prediction threshold \predth, we can get perfect performance on ZT cases (as nothing will ever be predicted). However, performance for ST and MT cases will drop. We find $\predth=0.1$ to be the optimal value. }
\label{fig:results_val_bceloss_iou}
\end{figure*}

\section{Analysis of matching strategies}
\vspace{-6pt}
\label{sec:supp:analysis}
\label{sec:analysis_multi3d_suppl}
We study the effect of different matching strategies (\emph{Hungarian} vs \emph{All}) on the performance of the D3Net and \OURMETHODCLIP on the \PROJECTNAME task. We also vary the matching IoU thresholds \ioutrainth (0.25 vs 0.50) and prediction confidence thresholds \predth (from 0.0 to 0.4).  We plot the F1 at IoU of 0.5 for the different variants using the 5-scenarios we established (\Cref{fig:results_val_bceloss_iou}).

\vspace{-2pt}
\xhdr{5-scenario breakdown.}
We identify our 5 scenarios (ZT w/ D, ZT w/o D, ST w/ D, ST w/o D and MT) according to the nyu40 %
semantic label set. Note that ZT metrics are special cases which report $F1{=}1$ if the model predicts nothing and $F1{=}0$ if the model predicts too many. 

\vspace{-2pt}
\xhdr{Prediction threshold.}
We further study different \predth used to filter out model outputs. \Cref{fig:results_val_bceloss_iou} shows that all models achieve the best performance at $\predth=0.1$.

\vspace{-2pt}
\xhdr{Matching strategies.}
We compare the two matching strategies (\emph{Hungarian} vs \emph{All}) that we used to set up positive and negative instances between the GT bounding boxes and proposed bounding boxes for calculating the reference loss $L_\text{ref}$. We compare results between \OURMETHODCLIP and D3Net~\cite{chen2022d3net}. \Cref{fig:results_val_bceloss_iou} shows that \emph{Hungarian} (darker lines) outperforms \emph{All} (lighter lines) on both methods, especially when \ioutrainth is small (e.g. $0.25$), since \emph{Hungarian} guarantees an optimal one-to-one matching. When \ioutrainth is larger (e.g. $0.5$), the gap caused by these two strategies gradually narrows. For D3Net, the two matching strategies do not exhibit noticeable differences.  We suspect this is due to a less noisy detector and that \emph{Hungarian} matching is effective when proposals are noisy.

\begin{figure*}
  \centering
  \includegraphics[width=\linewidth]{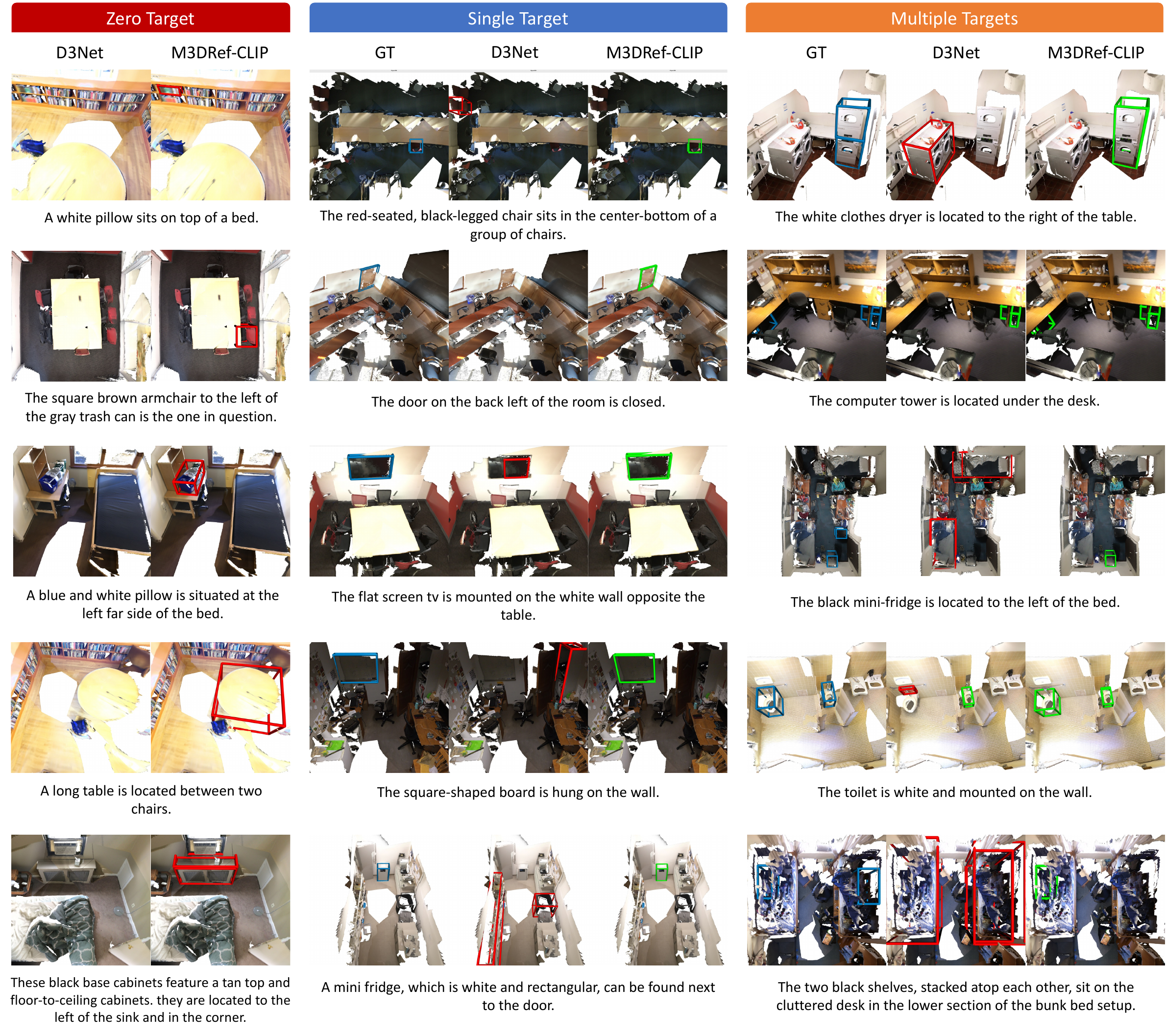}
  \vspace{-18pt}
  \caption{Qualitative results of D3Net~\cite{chen2022d3net} versus \OURMETHODCLIP on \PROJECTNAME using predicted boxes. Blue boxes indicate GT, green boxes are true positives with IoU threshold $\predth>0.5$. Red boxes are false positives.}
  \label{fig:qualitative_clip_suppl}
\end{figure*}

\section{Qualitative results on \PROJECTNAME}
\vspace{-6pt}
\label{sec:supp:qual_results}
In \Cref{fig:qualitative_clip_suppl}, we show qualitative examples of outputs from D3Net~\cite{chen2022d3net} and \OURMETHODCLIP for zero, single, and multiple targets.  In the Zero Target case (column 1), M3DRef-CLIP tends to predict false positives.  In the Single Target case (column 2), \OURMETHODCLIP has more accurate bounding boxes.  For Multiple Targets case (column 3), \OURMETHODCLIP identifies small objects accurately while D3Net has false detections of large objects (row 3,5).

\end{document}